\documentclass{article}

 \usepackage{tcolorbox} 
 \tcbuselibrary{skins, theorems}
 
\usepackage{microtype}
\usepackage{graphicx}
\usepackage{booktabs} 

\usepackage{amsmath}
\usepackage{amssymb}
\usepackage{mathtools}
\usepackage{amsthm}

\usepackage{enumitem} 
\usepackage{mdframed}
\usepackage{subfig} 
\usepackage{siunitx}

\usepackage[utf8]{inputenc} 
\usepackage[T1]{fontenc}    
\usepackage{nicefrac}       

\newcommand{\Figref}[1]{Figure~\ref{#1}}  
\newcommand{\figref}[1]{Figure~\ref{#1}}    

\newcommand{\tabref}[1]{Table~\ref{#1}}

\def\eqref#1{(\ref{#1})}

\newcommand{\secref}[1]{Sec.~\ref{#1}} 

\usepackage{xspace}
\makeatletter
\DeclareRobustCommand\onedot{\futurelet\@let@token\@onedot}
\def\@onedot{\ifx\@let@token.\else.\null\fi\xspace}
\makeatother

\newcommand{\eg}{e.g\onedot}
\newcommand{\ie}{i.e\onedot}

\usepackage{xr-hyper}
\makeatletter
\newcommand*{\addFileDependency}[1]{
  \typeout{(#1)}
  \@addtofilelist{#1}
  \IfFileExists{#1}{}{\typeout{No file #1.}}
}
\makeatother

\usepackage{xcolor}
\definecolor{ourblue}{rgb}{0.368,0.507,0.71}
\definecolor{ourorange}{rgb}{0.881,0.611,0.142}
\definecolor{ourgreen}{rgb}{0.56,0.692,0.195}
\definecolor{ourred}{rgb}{0.923,0.386,0.209}
\definecolor{ourviolet}{rgb}{0.528,0.471,0.701}
\definecolor{ourbrown}{rgb}{0.772,0.432,0.102}
\definecolor{ourlightblue}{rgb}{0.364,0.619,0.782}
\definecolor{ourdarkgreen}{rgb}{0.572,0.586,0.}

\definecolor{ourcyan2}{rgb}{0.125,0.722,0.804}
\definecolor{ourred2}{rgb}{0.863,0.184,0.047}
\definecolor{ouryellow2}{cmyk}{0,0.16,1.0,0.07}
\definecolor{ourviolet2}{cmyk}{0.55,0.56,0,0.47}
\definecolor{ourorange2}{cmyk}{0,0.46,0.89,0.11}

\theoremstyle{plain}

\theoremstyle{definition}

\theoremstyle{remark}

\newcommand*{\SO}{\mathrm{SO}}
\newcommand*{\SVD}{\ensuremath{\mathbb{R}^9}+SVD\xspace}
\newcommand*{\GSO}{\ensuremath{\mathbb{R}^6}+GSO\xspace}
\newcommand*{\htrue}{\ensuremath{h^*}}
\newcommand*{\hreal}{\ensuremath{\tilde h}}
\newcommand*{\pmm}{\ensuremath{\pm}}

\newcommand{\fplus}{\textcolor{teal}{\textbf{+}} \:}
\newcommand{\fminus}{\textcolor{purple}{\,\textbf{-}} \,\:}

\newcommand{\Real}{\ensuremath{\mathbb R}}        
\newcommand{\T}{\ensuremath{\top}}                

\graphicspath{{icml2024/images/}{images/}}

\usepackage[accepted]{icml2024}
\usepackage{hyperref}
\usepackage{multirow}

\hypersetup{
	colorlinks=true,       
	linkcolor=ourblue,        
	citecolor=ourdarkgreen,        
	filecolor=ourblue,     
	urlcolor=ourviolet
}

\usepackage[capitalize,noabbrev]{cleveref}
\usepackage{comment}

\definecolor{UntraversableRed}{RGB}{192, 26, 38}

\usepackage[textsize=tiny]{todonotes}

\icmltitlerunning{Learning with 3D rotations: a hitchhiker's guide to SO(3)}

\begin{document}
\twocolumn[
\icmltitle{Learning with 3D rotations, a hitchhiker's guide to SO(3)}




\begin{icmlauthorlist}
\icmlauthor{A. Ren\'e Geist}{1,2} 
\icmlauthor{Jonas Frey}{1,3}
\icmlauthor{Mikel Zhobro}{1,2}
\icmlauthor{Anna Levina}{4}
\icmlauthor{Georg Martius}{1,2}
\end{icmlauthorlist}

\icmlaffiliation{1}{Max Planck Institute for Intelligent Systems, Autonomous learning group, Germany}
\icmlaffiliation{2}{University of Tübingen, Distributed intelligence lab, Germany}
\icmlaffiliation{3}{Swiss Federal Institute of Technology (ETH Zurich), Robotic systems lab, Switzerland}
\icmlaffiliation{4}{University of Tübingen, Self-organization of neuronal networks group, Germany}

\icmlcorrespondingauthor{A. René Geist}{andreas-rene.geist@wsii.uni-tuebingen.de}

\icmlkeywords{Machine Learning, Rotations}

\vskip 0.3in
]



\printAffiliationsAndNotice{} 

\begin{abstract}
Many settings in machine learning require the selection of a rotation representation. 
However, choosing a suitable representation from the many available options is challenging.
This paper acts as a survey and guide through rotation representations. We walk through their properties that harm or benefit deep learning with gradient-based optimization.
By consolidating insights from rotation-based learning, we provide a comprehensive overview of learning functions with rotation representations.
We provide guidance on selecting representations based on whether rotations are in the model's input or output and whether the data primarily comprises small angles.

The project code is available at:
\newline \href{https://github.com/martius-lab/hitchhiking-rotations}{github.com/martius-lab/hitchhiking-rotations}
\end{abstract}

\section{Introduction}
For centuries, researchers explored methods to describe the rotation of Cartesian coordinate systems in three-dimensional Euclidean space. \citet{euler1765mouvement} demonstrated the necessity of at least three parameters for parameterizing 3D rotations through rotation matrices $R \in \text{SO}(3)$ as exemplified by Euler angles.
Subsequent discoveries led to the description of rotation through exponential coordinates and quaternions, which found widespread adoption in many fields such as control theory, robotics, and computer animations. The choice of a rotation representation for a specific application depends on its properties, including human-interpretability, computational costs, and dimensionality. In recent years, the representation's impact on learning functions from data has become another significant factor.

\begin{figure}[t]
    \centering
    \includegraphics[width=\linewidth]{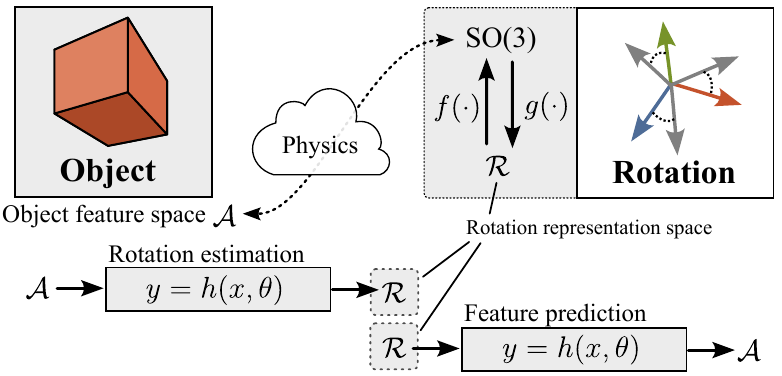}
    \vspace{-0.3cm}
    \caption{Overview on learning with rotations. A neural network learns a function from a feature space $\mathcal{A}$ to a rotation representation space $\mathcal{R}$ or vice versa. When learning with rotations, the properties of $f\,:\,\mathcal{R} \rightarrow \SO(3)$ and $g\,:\,\SO(3)\rightarrow \mathcal{R}$ affect training.}
    \label{fig:learning_overview}
    \vspace{-0.3cm}
\end{figure}

In machine learning, we want to choose the representation of our data that yields the best test accuracy. However, recent works \cite{zhou2019continuity, bregier2021deep, levinson2020analysis} suggest that rotation representations with four or less dimensions do not facilitate sample-efficient learning. To further complicate matters, empirical findings by \citet{pepe2022learning} suggest that geometric algebra representations may rival the high-dimensional representations proposed by \citet{zhou2019continuity, levinson2020analysis}.
Despite many recent works discussing the role of rotations in machine learning, we did not encounter a work that provides a comprehensive overview on the question:
\begin{quote}
\centering
    \emph{What representation of $\,\SO(3)$ is suitable for neural network regression with gradient-based optimization?}
\end{quote}
Besides considering the case in which rotations are in the models' output as discussed in \citet{zhou2019continuity, bregier2021deep, levinson2020analysis}, we also detail how one may learn functions from rotations being in the input.
We discuss common rotation representations, using tricks such as wrapping angle coordinates into sinusoidal transformations, problems arising from double-cover, and how the choice of metric affects training. 
We find that high-dimensional representations should be the preferred default choice from a theoretical as well as empirical perspective. 
As it turns out, many aspects of a rotation representation influence training while also vice versa, the setup of machine learning problem influences the choice of rotation representation.

\begin{table*}[t]
\caption{Overview of selected rotation representations with some of their properties.
Due to singularities arising from angle coordinates and double cover (explained below), rotation representations with three and four dimensions have a discontinuous  map $g(R): \SO(3)\to \mathcal{R}$.}\label{tab:representations}
\centering \vspace{.3em}
\footnotesize
\begin{tabular}{ll@{}c@{\hspace{0.5ex}}c@{\hspace{3.5ex}}lccc}
\toprule
\textbf{2D/3D} & \textbf{Representation}  & \textbf{Notation} & \textbf{Dim}  &  \hspace{-2ex}\textbf{Domain} & $g(R)$ \textbf{cont.} & \textbf{Uses Angles} & \textbf{Double cover} \\
\midrule
\multicolumn{1}{l|}{\multirow{2}{*}{$\SO(2)$}} & Angle & $\alpha$ & 1 & $\mathbb{R}^1$ & \fminus &  yes & no \\
\multicolumn{1}{l|}{}  & Sine and cosine of the angle & $[\cos(\alpha), \sin(\alpha)]$ & 2 & $\mathbb{R}^2$ & \fplus &  no & no \\
\midrule
\multicolumn{1}{l|}{\multirow{6}{*}{$\SO(3)$}}  & Euler parameters & Euler & 3 & $\mathbb{R}^3$ & \fminus & yes & yes \\
\multicolumn{1}{l|}{}  & Exponential coordinates / Bi-Vectors & Exp & 3 & $\mathbb{R}^3$ & \fminus & no & yes \\
\multicolumn{1}{l|}{}  & Unit-quaternions /  Rotors & Quat &  4 & $S^3$ & \fminus & no & yes \\
\multicolumn{1}{l|}{}  & Axis-angle & Axis-angle& 4 & $\mathbb{R}^4$ & \fminus & yes & yes \\
\multicolumn{1}{l|}{}  & \ensuremath{\mathbb{R}^6}+Gram-Schmidt orthonormalization & \GSO & 6 & $\mathbb{R}^{3 \times 2}$ & \fplus & no & no \\
\multicolumn{1}{l|}{}  & \ensuremath{\mathbb{R}^9}+singular value decomposition & \SVD & 9 & $\mathbb{R}^{3 \times 3}$ & \fplus & no & no \\
\bottomrule
\end{tabular}
\normalsize
\end{table*}

\subsection{Problem setting}
In this work, we consider gradient-based supervised neural network regression.
That is, given a data set $\mathcal{D}=\{x^{(i)},y^{(i)}\}_{i=1}^N$ of $N$ inputs $x \in \mathcal{X}$ and outputs $y\in \mathcal{Y}$, find the parameters $\theta$ of the neural network $h: \mathcal X \to \mathcal Y$
\begin{equation} \label{eq:model}
    y = h(x; \theta)
\end{equation}
that minimize the loss function
\begin{equation}
    L(\mathcal D, \theta) = \sum_{x,y\in \mathcal D} d(y,h(x,\theta)) \label{eq:loss}
\end{equation}
 using the parameter gradient $\nabla_{\theta} L$.
The loss measures the average distances $d$ of predictions to target values.
For simplicity, we assume that there exists a deterministic target function $y=\htrue(x)$ that generated the data.

\paragraph{Representing rotations}
To work with rotations, they have to be suitably represented. In analogy to \citet{zhou2019continuity}, we say a vector $r \in \mathcal{R} \subseteq \mathbb{R}^d$ is a rotation representation, if there exist two functions $f: \mathcal{R} \to \SO(N)$ and $g: \SO(N) \to \mathcal{R}$ such that $f(g(R))=R$ that is $f$ is a left inverse of $g$.
Note that depending on the choice of the representation the topology and dimensionality $d$ of $\mathcal{R}$ may vary. \tabref{tab:representations} lits common representations that differ in their dimensionality and topology.

\paragraph{Learning scenarios}
To analyze the consequences of different representations on learning, we consider the cases where rotations occur in the input ($\mathcal{R} \subseteq \mathcal{X}$) and where they occur in the output ($\mathcal{R} \subseteq \mathcal{Y}$) of our regression problem.
As depicted in \figref{fig:learning_overview}, we denote the features of an entity of interest by features $a \in \mathcal{A}$.
This could be, for instance, a camera image, a point cloud, or adhesive forces acting between two molecules.
For simplicity, we ignore possible other components and consider the two pure cases of:

\begin{enumerate}[label=\roman*),nosep]
    \item \emph{feature prediction} where $a = h(r)$ \\with $\mathcal{X}=\mathcal{R}$ and $\mathcal{Y}=\mathcal{A}$,
    \item \emph{rotation estimation} where $r = h(a)$ \\with $\mathcal{X}=\mathcal{A}$ and $\mathcal{Y}=\mathcal{R}$,
\end{enumerate}
In feature prediction, we are interested in learning a function from $r$ (and other object variables) to a particular object property $a$, for example, rendering an object from a particular direction or predicting dynamics.
In rotation estimation, a map from a high-dimensional representation $a$ to $r$ is learned, for example, pose estimation from images.

\section{Representations of rotation} \label{sec:rotation-representations}
In this section, we describe common rotation representations as detailed in \tabref{tab:representations} and examine their properties. An overview is provided in \tabref{tab:representations}.

\begin{figure}
    \centering
     \includegraphics[width=0.28\linewidth]{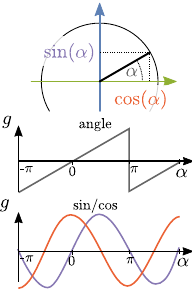}\qquad
    \includegraphics[width=0.6\linewidth, trim={0 0 0.2 0}, clip]{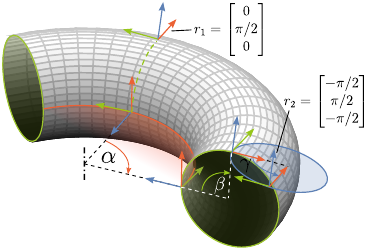}
    \vspace{-0.3cm}
    \caption{Left: Representations for SO(2): angle and sin/cos with respective $g$ functions. Right: Euler angles representation for SO(3). In Euler angles a frame can be visualized as three consecutive SO(2) rotations along the surface of a torus.
    }
    \label{fig:2d-and-euler-torus}
\end{figure}

\paragraph{SO(2) representations}
Rotation angles seem to be the most natural representation. A 2D rotation is fully described by a single angle $\alpha$. However, because the angle is a 1D quantity, there is a jump in the function $g$, as shown on the left in \figref{fig:2d-and-euler-torus}.
If the rotation is represented by $(\cos(\alpha), \sin(\alpha))$, such a discontinuity does not arise.

\paragraph{Euler angles}
In 3D we need at least three angles $\alpha,\beta,\gamma\in [-\pi,\pi)$ (Euler angles) to describe a rotation.
Rotation matrices can be composed through a sequence of elementary rotations $R(\alpha,\beta,\gamma) = R_3(\gamma)R_2(\beta)R_1(\alpha):=f(r)$ by these angles.
Discontinuities arise when bounding the range of angles (see above).
In addition, for Euler angles, the same point in $\SO(3)$ can be described by different representation vectors. For example, in \figref{fig:2d-and-euler-torus} the points $r_1=[0,\pi/2,0]$ and $r_2=[-\pi/2, \pi/2, -\pi/2]$ perform the same rotation.

Due to these and other reasons, studies on learning with rotations, including \citet{huynh2009metrics, zhou2019continuity, bregier2021deep, pepe2022learning}, uniformly discourage the use of Euler angles in learning with 3D rotations.

\begin{figure}[t]
    \vspace{-1mm}
    \centering
    \includegraphics[width=0.75\linewidth]{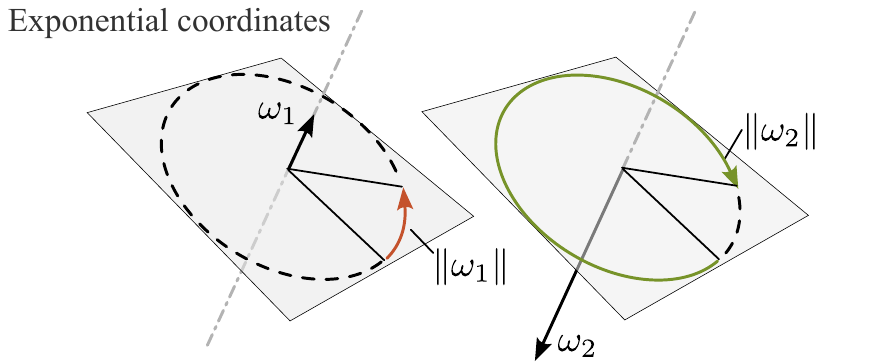}
    \vspace{-3mm}
    \par\noindent\rule{4cm}{0.4pt}\\
    \vspace{1mm}
    \includegraphics[width=0.75\linewidth]{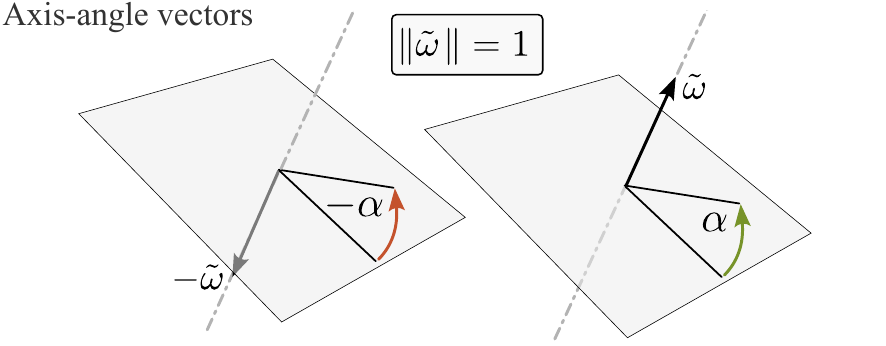}
    \vspace{-2mm}
    \caption{Exponential coordinates (Exp) and axis-angle representation and their double cover property.
    \textbf{Top}: Exp. coord.: rotation around $\omega$ by angle $\|\omega\|$.  The vector $\omega_1= \alpha_1 \tilde \omega_1 \in \mathbb{R}^3$ describes the same rotation as $\omega_2=(\|\omega_1\|-2\pi)\omega_1/\|\omega_1\|$.
    \textbf{Bottom:} Axis-angles explicitly represent axis and angle: $\tilde \omega \in \mathcal{S}^2$, $\alpha \in \mathbb{R}$.
    The vector $[\tilde \omega, \alpha]$ describes the same rotation as $[-\tilde \omega, -\alpha]$. }
    \label{fig:axis-angle-exp}
    \vspace{-4mm}
\end{figure}

\paragraph{Exponential coordinates}
Each rotation in 3D can also be expressed by a rotation axis $\omega \in\mathbb{R}^3$ and an angle. One possible representation is thus to use the length of the vector $\|w\|$ to encode the angle.
The identity rotation corresponds to the origin of $\Real^3$ (zero vector) and spheres of radius $2n\pi$ with $n\in \mathbb{N}^+$.
This representation is called
\emph{exponential coordinates} because $\omega$ can be written as a $3\times3$ skew-symmetric matrix. The matrix exponential of that yields the desired rotation matrix. Alternative names are ``rotation vectors'', ``Rodrigues parameters'', or ``angular velocity''.
\Figref{fig:axis-angle-exp} shows the concept. It also illustrates that the same rotation can be expressed by two vectors (inside the sphere of radius $2\pi$) as $f(\omega)=f\big((\|\omega\|-2\pi)\frac{\omega}{\|\omega\|}\big)$ (double cover).

\paragraph{Axis angle and Quaternions}
Instead of tying the length of the axis vector to the rotation, we can separately represent them as $r = (\tilde\omega, \alpha)\in\Real^4$ with $\|\tilde\omega\|=1$, logically denoted as \emph{axis angle}. \cref{fig:axis-angle-exp} illustrates the same double cover property as exponential coordinates.  We can transform exponential coordinates to axis angles and then the axis angle vector to $\SO(3)$ via Rodrigues' rotation formula $f(r):=\mathbb I + \sin(\alpha) [\tilde \omega]_{\times} + (1-\cos(\alpha))[\tilde \omega]_{\times})^2$ whereas the map from $\SO(3)$ to exponential coordinates is given by the matrix exponential $g(R):=\exp(R)$.

Similar representations are \emph{quaternions}, which extend the concept of complex numbers to higher dimensions. For the purpose of rotations, we consider unit quaternions that relate to the axis angle representation $(\tilde\omega, \alpha)$ as $r = q = (w,x,y,z) \in \mathcal S^3$, \ie $\|q\|=1$ and $w= \cos(\alpha/2)$ and $(x,y,z) = \sin(\alpha/2) \tilde\omega$ \cite{grassia1998practical}.
Unit quaternions double cover $\SO(3)$ such that $f(q)=f(-q)$.

\paragraph{$\Real^6$ + Gram-Schmidt orthonormalization (GSO)}
A representation that is closer to the actual rotation matrix is $r = (\nu_1,\nu_2) \in \Real^{3 \times 2}$.
A rotation matrix can be obtained by using Gram-Schmidt orthonormalization (GSO) to ``complete'' a Cartesian frame \cite{zhou2019continuity}.
As the columns of a rotation matrix are unit length and orthogonal to each other, this method yields a rotation matrix $R=f(r) = \mathrm{GSO}(\nu_1, \nu_2)$ whereas $g(R):=\mathrm{diag}(1,1,0) R$. The simple idea of GSO is to first normalize $\nu_1$, subtract from $\nu_2$ its component that is colinear with $\nu_1$ to obtain $\nu_2^{\perp}$, then after normalizing $\nu_2^{\perp}$, we obtain $\nu_3$ as the cross product of $\nu_1$ and $\nu_2^{\perp}$.
GSO straightforwardly extends to $\SO(n)$ \citep[p.\,120]{macdonald2010linear}.

\paragraph{$\Real^9$ + singular value decomposition (SVD)}
Rotations can be directly parameterized by a $3 \times 3$ matrix which is then projected to $SO(3)$ using
the singular value decomposition (SVD). Given $r = M\in\mathbb{R}^{3 \times 3}$ SVD decomposes the matrix into $M=U \Sigma V^T$ where $U,V \in \mathbb{R}^{3 \times 3}$ are rotations or reflections and $\Sigma=\text{diag}(\sigma_1, \sigma_2, \sigma_3)$ is a diagonal matrix with the singular values $\sigma_i$ denoting scaling parameters.
Now, the projection is achieved by
\begin{equation}
    f(r):= \text{SVD}^+(M) = U \, \mathrm{diag}(1, 1, \det(UV^T)) \,V^T,
\end{equation}
where $\det(UV^T)$ ensures that $\det(\text{SVD}^+(M)) = 1$ \cite{levinson2020analysis}. As we shall see later, this operation finds the rotation matrix with the least-squares distance to $M$. The function from $\SO(3)$ to $\mathcal{R}$ is simply $g(R):=R$.

\section{Measuring distances between rotations} \label{sec:metrics}
Supervised regression requires us to measure distances in $\mathcal{Y}$ (\cref{eq:loss}).
A proper distance metric $d(y_1,y_2)$ is nonnegative, only zero if $y_1=y_2$ (identity), symmetric, and satisfies the triangular inequality: $d(y_1,y_2)\leq d(y_1,y_3)+d(y_3,y_1)$, yet, when employed as a loss function in machine learning some of these properties can be lifted.
When it comes to the estimation of rotation, we need to measure the distances either in $SO(3)$ or in $\mathcal{R}$.
In the following discussion, we give a brief overview and geometric understanding of the frequently encountered metrics in rotation learning. A comprehensive overview of the mathematical properties of those metrics is given in \citet{huynh2009metrics, hartley2013rotation, alvarez2023loss}.

\begin{figure}
    \centering
    \includegraphics[width=0.8\linewidth]{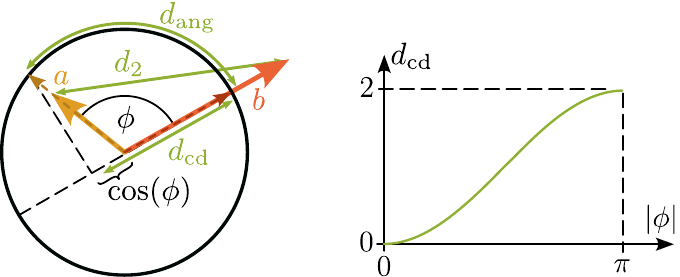}
    \vspace{-3mm}
    \caption{Geometric illustration of distance metrics $d(a,b)$ between vectors $a$ and $b$. Cosine distance ($d_\textrm{cd}$) and angular distance ($d_\textrm{ang}$) ignore the vectors' lengths.}
    \label{fig:vector_metrics}
\end{figure}
\paragraph{Metrics on Euclidean vector representations}
We start with general measures (metric and non-metric) that are used to evaluate the similarity between vectors.
The Euclidean distance measures the length of the difference vector
$    d_{2}(a,b) = \|a-b\|_2,$
where the $L^2$ norm of $a$ reads $\|a\|_2:=\|a\|=\sqrt{a \cdot a}$.
Using $\cos(\phi)=\frac{a \cdot b}{\|a\| \|b\|}$, we can define
the cosine distance and angular distances as:
\begin{equation} \label{eq:cosine_distance}
d_{{cd}}(a,b) = 1 - \cos(\phi)\text{,\  and \ } d_{\mathrm{ang}}= \arccos(\cos(\phi)).
\end{equation}
The angular distance ($\in [0,\pi]$) measures the geodesic distance on a sphere $\mathcal{S}^n$, the cosine distance ($\in [0,2]$) measures a projection distance of the two (normalized) vectors (\figref{fig:vector_metrics}). Note that cosine distance and angular distance are pseudo-metrics as they violate identity (ignoring the length of the vectors), and cosine distance does not satisfy the triangular inequality, which does not cause problems in practice.
 Both metrics yield the same ordering of vectors, i.e., for any set of vectors $v_1,\ldots,v_M$, the ordering from closest to furthest from vector $u$ would be the same for both measures.

\paragraph{Metrics that pick distances}
In \secref{sec:rotation-representations}, we encounter the problem of double cover, \ie the same rotation in $SO(3)$ is expressed by  two different points in the representation space (\figref{fig:2d-and-euler-torus}).
For the case of quaternions, we have $f(q)=f(-q)$ such that $d_2(q,-q)\neq 0$ even though the corresponding rotation matrices are identical. An attempt to circumvent this problem is to pick the shortest metric between the quaternions and their negative complement:
\begin{align}
    d_{q,I} &= \min(\|q_1-q_2\|, \|q_1+q_2\|), \text{ or } \label{eq:distance-picking-1}\\
    d_{q,II} &= 1 - |q_1 \cdot q_2|. \label{eq:distance-picking-2}
\end{align}
For Euler angles $\alpha, \beta, \gamma \in \mathbb{R}$, \citet{huynh2009metrics} proposed
\begin{equation} \label{eq:distance-picking-3}
    d_{\mathrm{e}}(r_1, r_2)=\sqrt{ d(\alpha_1, \alpha_2)^2 + d(\beta_1, \beta_2)^2 + d(\gamma_1, \gamma_2)^2 }
\end{equation}
with $d(a,b)=\min(|a-b|,2\pi-|a-b|)$ and $\beta \in [-\pi/2,\pi/2]$ to avoid that two sets of Euler angles represent the same element of $\SO(3)$.
Note that \eqref{eq:distance-picking-1}, \eqref{eq:distance-picking-2}, and \eqref{eq:distance-picking-3} are pseudo-metrics on $\mathcal{R}$ but act as metrics on $\SO(3)$.

\paragraph{Metrics on rotation matrices}
The distance between rotation matrices is typically measured using the Frobenius norm, also called Schurnorm.
The matrix Frobenius norm of a square matrix $R = [v_1, v_2, \dots, v_n] \in \mathbb{R}^{n \times n}$ reads
\begin{equation} \label{eq:frobenius-norm}
\|R\|_{\text{F}} = \sqrt{ \sum_i^n \sum_j^n R_{i,j}^2 }  = \sqrt{ \sum_i^n \|v_i\|^2 }=\|\mathrm{vec}(R)\|. 
\end{equation}
Recall that the columns of a rotation matrix $v_i$ are the basis vectors of the rotated coordinate frame.
By inserting the difference between two rotations into the Frobenius norm \eqref{eq:frobenius-norm} one obtains the  Chordal distance as
\begin{align}
    d_c(R_1, R_2) &= \|R_1-R_2\|_{\text{F}}= \sqrt{ \sum_i^n \|v_{1,i} - v_{2,i}\|^2 }, \label{eq:F_m1}
\end{align}
which is an often used metric due to its numerical stability
and computational efficiency. For $R_1,R_2 \in \SO(3)$ we have $d_c(R_1, R_2) \in [0,2\sqrt{2}]$. Due to advantageous convexity properties, the squared Chordal distance $d_c(R_1, R_2)^2$ is often used in practice \cite{hartley2013rotation}. In turn, the mean squared error between two set of rotations $\{R_{1,1},...,R_{1,N}\}$ and $\{R_{2,1},...,R_{2,N}\}$ reads $MSE=\sum_{i=1}^N d_c(R_{1,i},R_{2,i})^2/N = \sum_{i=1}^N\|\mathrm{vec}(R_{1,i}-R_{2,i})\|^2/N$.
The geodesic distance can be obtained using multiplication $R_1 R_2^\T$ (which would be $I$ if $R_1 = R_2$) as
\begin{equation}
    d_{\phi}(R_1, R_2) = \arccos{\frac{\text{tr}(R_1R_2^T) - 1}{2}}.
\end{equation}

\cref{fig:distance_gradients} in the appendix shows the loss function gradients of several of the discussed functions in $\mathbb{R}^2$. Depending on which loss is used the gradients are oriented differently relatively to the unit circle $\mathcal{S}^1$.

\section{Rotation representations affect learnability}
\begin{figure}[h]
    \centering
    \includegraphics[width=0.9\linewidth]{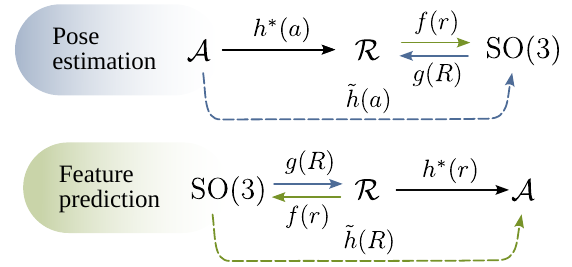}
    \vspace{-2mm}
    \caption{The target function $\htrue(x)$ is the composition between $\tilde h(x)$ and the functions $g(R)$ / $f(r)$.
   }
    \label{fig:maps}
\end{figure}

We analyze now how rotation representations affect the properties of the function $\htrue(x)$ which we want to learn with \cref{eq:model}.
For that, consider the true mapping $\hreal$ from the feature space $\mathcal A$ to $\SO(3)$ or vice versa.
An interesting observation is depicted in \figref{fig:maps}: the target function $\htrue$ is actually the composition of the functions:
\begin{enumerate}[nosep,label=(\roman*)]
\item For rotation estimation, $\htrue = g \circ \hreal$; \label{enum:rot-estimation}
\item For feature prediction, $\htrue = \hreal \circ f$. \label{enum:feature-pred}
\end{enumerate}

With this realization, it is evident that for rotation estimation \ref{enum:rot-estimation} discontinuities in $g$ or double representations can translate to $\htrue$. For successful gradient-based learning, however, we want $\htrue$ to exhibit some notion of \textbf{continuity}, as those functions are easier to learn \cite{xu2004essential, xu2005simultaneous, llanas2008constructive}.
The \emph{pre-images connectivity constraint} on $f$, guarantees that there always exists a smooth function that allows to properly interpolate between training representations \cite{bregier2021deep}.
For feature prediction \ref{enum:feature-pred}, $f$ is continuous for all representations and does not pose a problem with regards to continuity. Albeit, we observe a problem from \textbf{disconnectedness}, which we discuss below.

\subsection{Why does rotation estimation need high-dimensional rotation parameterization?} \label{sec:4.1}
To quantify the continuity of a function, we use the concept of Lipschitz continuity
 that bounds the maximal slope to a constant $\mathcal{L}$ (which would also lead to continuity and require that infinitesimal changes in the function's input result in infinitesimal changes in its output).
The mapping function $g(R)$ is Lipschitz continuous if there exist $\mathcal{L}\geq 0$ such that for any $R_1, R_2 \in \SO(3)$ and $r_1 = g(R_1), r_2 = g(R_2), r_1,r_2\in \mathcal{R}$,
\begin{equation}
\frac{\|r_1 - r_2\|_2}{\|R_1 - R_2\|_{\mathrm{F}}} \leq \mathcal{L}.
\end{equation}

\begin{figure}[t]
    \centering
    \includegraphics[width=1.\columnwidth]{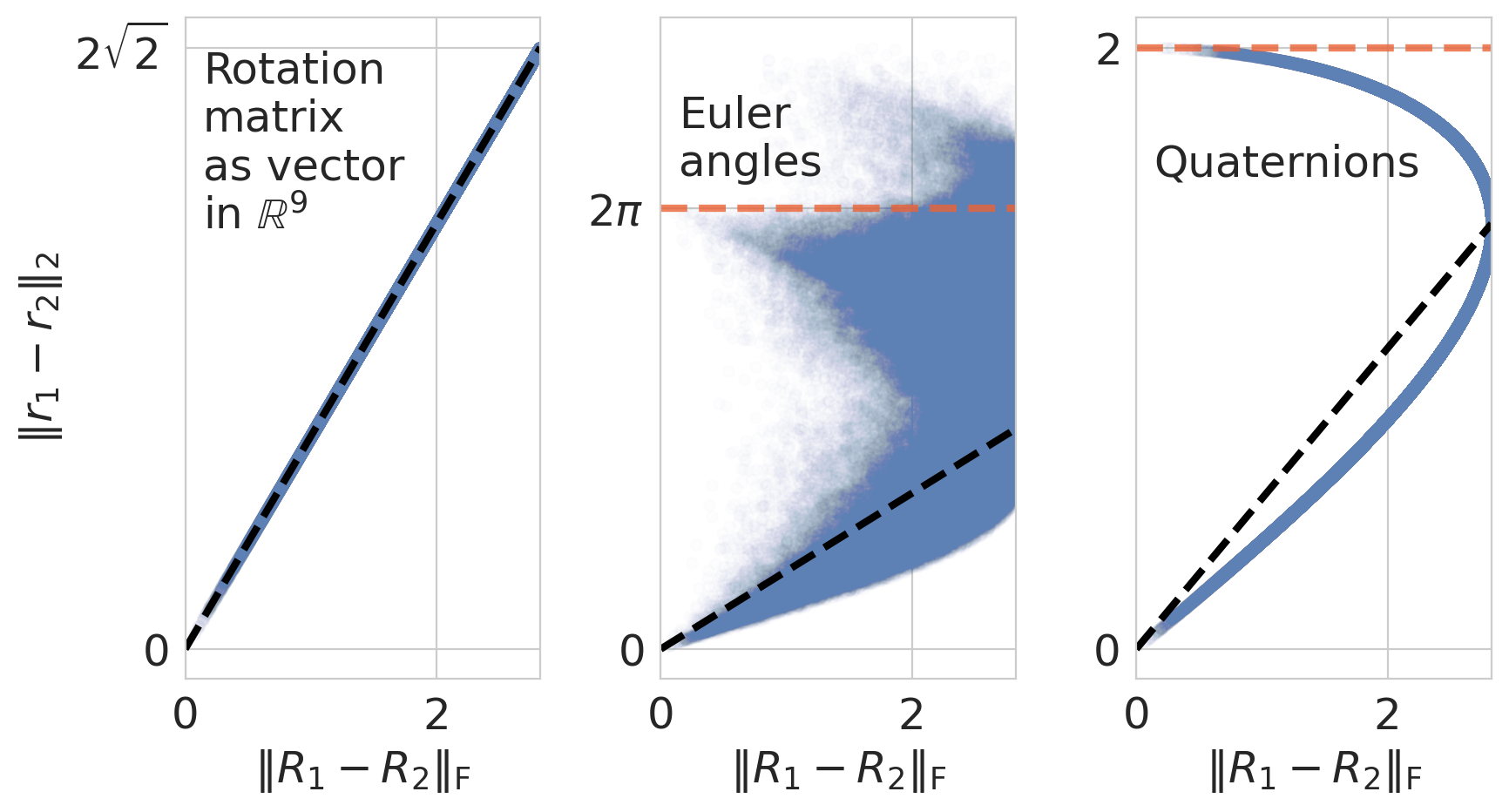}%
    \vspace*{-.5em}
    \caption{Chordal distance between randomly sampled $R_1,R_2\in\SO(3)$ and $L^2$ norm between corresponding rotation representations $r_1=g(R_1),r_2=g(R_2)$. The full width of $\mathcal{R}$ is marked by red lines. Ideally, the ratios between the distances, aka the Lipschitz constant of $g(R)$, is close to the black lines whose slope amounts to the ratio between the minimum width of $\mathcal{R}$ and $\SO(3)$.
    }
    \label{fig:lipschitz_constants}
\end{figure}

\paragraph{Discontinuity of $g$ for low-dimensional representations.}
$SO(3)$ is not homeomorphic to any subset of 4D Euclidean space; thus, for rotation representations with four or fewer dimensions $g(R)$ must be discontinuous (therefore,  not Lipschitz continuous) \cite{ stuelpnagel1964parametrization, levinson2020analysis}.
\Cref{fig:lipschitz_constants} illustrates these discontinuities by randomly sampling $R_1,R_2\in\SO(3)$ and then mapping these representations to $r_1=g(R_1)$ and $r_2=g(R_2)$. For low-dimensional representations, small distances in $\SO(3)$ can lead to close to maximum distances in $\|r_1-r_2\|$.

\paragraph{Double Cover}
A rotation representation doubly covers $\SO(3)$ if every element of $\SO(3)$ is represented by two different elements of $\mathcal{R}$.
Examples among the representations are unit-quaternions where $f(q)=f(-q)$,  exponential coordinates where $f(\omega)=f((\|\omega\|-2\pi)\frac{\omega}{\|\omega\|})$, and axis-angle vectors where $[\tilde \omega, \phi]=[-\tilde \omega, -\phi]$.
Double cover leads to a discontinuity in $g(R)$, and as a result, measuring the shortest distance between two representation vectors does not necessarily correspond to the shortest distance in $\SO(3)$.
How does this affect gradient-based machine learning?

\begin{figure}
    \centering
    \includegraphics[width=0.99\linewidth]{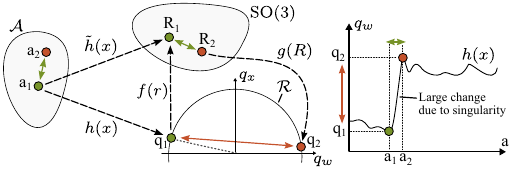}
    \vspace{-2mm}
    \caption{Rotation estimation suffers from double cover: We assume that a small change in the feature space $\mathcal{A}$ corresponds to a small change in $\SO(3)$ (green arrows). For some rotation representations $\mathcal{R}$ such as quaternions, small distances in $SO(3)$ are sometimes mapped to large distances in $\mathcal{R}$ (red arrows) yielding a discontinuous function $h$ (right).
   }
    \label{fig:map_illustration}
\end{figure}
We can resort to Lipschitz continuity to understand the problems arising from discontinuities. Assume we have two feature vectors $a_1,a_2 \in \mathcal{A}$ that are mapped by $\tilde h(x)$ to the nearby elements $R_1, R_2 \in \SO(3)$. The discontinuities in $g(R)$ imply that the distance between two representation vectors $r_1=g(R_1), r_2 = g (R_2) \in \mathbb{R}^d$ can still be large and  $\frac{\|r_2-r_1\|}{\|R_2-R_1\|} \xrightarrow[\|R_2-R_1\| \to 0]{} \infty$. Because of it, there must be points in feature space where $\htrue(x)= g \circ \hreal$ has Lipschitz constant $\mathcal{L}=\frac{\|r_2-r_1\|}{\|a_2-a_1\|}$ that blows up as $a_2-a_1$ approaches zero.
Naturally, this will cause the loss function's gradient $\nabla_{\theta} L$ to also blow up.
 \figref{fig:map_illustration} demonstrates this issue for the case of rotation estimation with quaternions yielding discontinuities that almost span across the full width of $\mathcal{R}$.

There seem to be two options: try to fix discontinuities or avoid them in the first place.
In the following, we briefly analyze different fixes proposed in the literature, and provide a recommendation for gradient-based learning.

\paragraph{Attempting to fix discontinuities via sin-cos angle coordinates:}
For any representation that contains angles, we can represent them as $[\cos(\cdot), \sin(\cdot)]$
to avoid the discontinuous jump at the boundaries of the interval $[-\pi, \pi)$.
While the above trick is useful for learning with representations of $\SO(2)$ and $\SO(2) \times \SO(2)$ (spherical coordinates), it does not remove discontinuities in $\SO(3)$ representations that arise from double cover.
For example, axis-angle still suffer from double cover $f(\omega, \alpha)=f(-\omega, -\alpha)$ even if $\alpha\in\SO(2)$ is replaced by $\cos(\alpha), \sin(\alpha)$.

\paragraph{Attempting to fix discontinuities via distance-picking or computing distances in $\SO(3)$:}
Distance-picking and measuring distances in $\SO(3)$ are common strategies found in literature related to rotation estimation. In distance picking, we choose the minimum distance between elements in $\mathcal{R}$ taking double cover into account \eg \cref{eq:distance-picking-1,eq:distance-picking-2}. Alternatively, we can directly compute distances in $\SO(3)$, \eg via $d_c(f(r_1), f(r_2))$, such that the gradients $\nabla_{y} d_c(f(y), f(r_2))$ point in directions in $\mathcal{R}$ that increase the metric in $\SO(3)$. 
However, from the previous discussion and the additional argument provided in \cref{sec:app-distance-picking}, we conclude the following key insight:

\begin{mdframed}[backgroundcolor=gray!20]
\textbf{Changing the loss does not fix discontinuities:} 
due to double cover, the target function maps similar features to vastly different rotation representations (one-to-many map). The subsequent issues arising in learning the target function are not fixed using distance picking or computing distances in $\SO(3)$.
\end{mdframed}

This observation is supported by experiments in \citet{alvarez2023loss} on visual odometry (estimate camera pose from images) and our experiments in \cref{sec:learning}.

\begin{figure}[t]
    \centering
    \includegraphics[width=.85\linewidth]{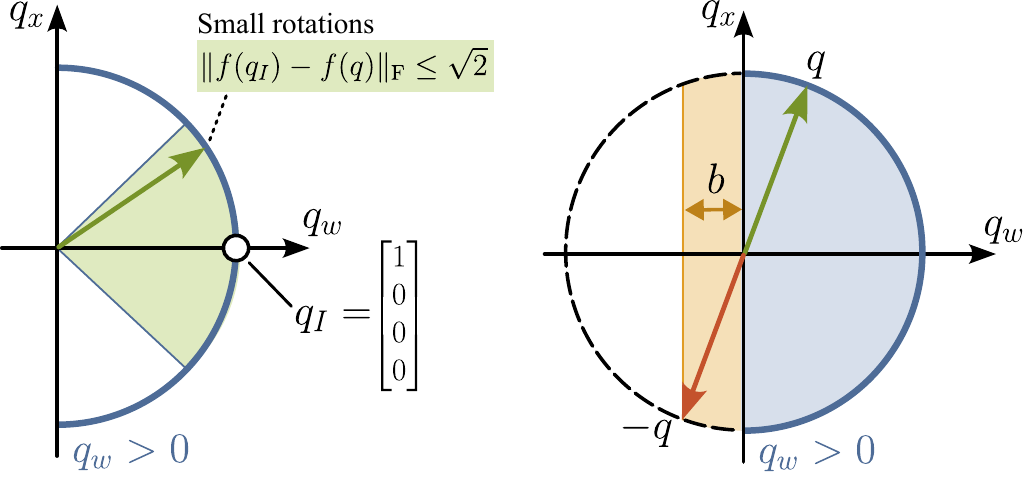}
    \vspace{-2mm}
    \caption{Half-space maps work for ``small'' rotation estimation (left) and data-augmentation for feature prediction (right). Left: As long as the rotation vectors are within the green cone, there are no discontinuities and the distances behave as expected. Right: To avoid underrepresented input regions, data-augmentation (orange) for vectors close to the half-space boundary can be used.
    }
    \label{fig:quaternion_doublecover}
\end{figure}

\paragraph{Attempting to fix discontinuities via half-space map:}
Another idea would be to constrain the representation in $\mathcal R$ to one \emph{half} such that double-cover is not present (except at the separation hyper-plane). For instance, for quaternions where $f(q)=f(-q)$, we simply flip all vectors with negative scalar dimension as shown in \figref{fig:quaternion_doublecover}. Also, the axis-angle representation and exponential coordinates allow splitting the space in two halves as shown in \figref{fig:axis-angle-exp}.

Again, the apparent fix has problems for vectors close to the separating hyperplane. Consider the case of quaternions as depicted in \figref{fig:map_illustration}.
Two vectors close to the boundary are far in $\mathcal{R}$ but close in $\SO(3)$, such that the Lipschitz-constant
of $g$ can grow unbounded.
Empirical evidence in \cref{sec:learning} and in \citet{saxena2009learning} on learning an object's orientation from images supports this.
Albeit, for small rotations where $\|\mathbb I-f(q)\|_{\mathrm{F}}\leq \sqrt{2}$ with the identity matrix $\mathbb I$ amounting to zero rotation as shown in \figref{fig:quaternion_doublecover} (left), a half-space map is a valid fix as no discontinuities are seen during training \cite{bregier2021deep}.
This case is particularly relevant in dynamics simulation, where changes in rotations remain small between timesteps. 

\paragraph{Circumventing discontinuities by using 6D or 9D representations:}
The obvious solution to the above problems is to use high-dimensional representations that have a continuous mapping $g$ (\cref{sec:rotation-representations}).
While \citet{bregier2021deep} provides a detailed analysis of such representations, we limit the discussion to the questions:
Why do these representations work well and why does \SVD outperform \GSO?

\begin{figure}
    \centering
    \includegraphics[width=.9\columnwidth]{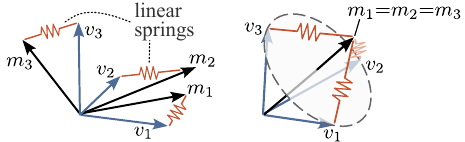}  
    \vspace{-2mm}
    \caption{Illustration of \SVD. Left: SVD yields the closest Cartesian frame $R\!=\![v_1, v_2, v_3]$ to the matrix $M\!=\![m_1, m_2, m_3]$. Think of springs that try to pull $v_i$ towards $m_i$.
    Right: Singularities arise when $\det(M)\!=\!0$: several $R$ minimize the potential energy.
    }
    \label{fig:svd}
    \vspace{-3mm}
\end{figure}

\paragraph{\SVD is smooth}
For the \SVD representation of rotation, \(f(r):=\text{SVD}^{+}(M)\) is smooth if $\text{det}(M) \neq 0$ with $r:=M$, as it solves the ``orthogonal Procrustes problem'' using the Frobenius norm \eqref{eq:frobenius-norm}:
\begin{align}
    \text{SVD}^+(M) &:=\underset{R\in \SO(3)}{\mathrm{arg\,min}}\: \|R-M\|_{\mathrm{F}}, \label{eq:procrustes-problem} \\
                    &:= \underset{R\in \SO(3)}{\mathrm{arg\,min}}\: \sum_{i=1}^3 \|v_i - m_i\|^2, \label{eq:potential_energy}
\end{align}
where $M = [m_1, m_2, m_3]$ and $R=[v_1, v_2, v_3]$.
This method projects $M$ onto the nearest rotation matrix in $\SO(3)$ which we visualize in \figref{fig:svd} as spring forces pulling $R$ towards minimum potential energy \eqref{eq:potential_energy}. The potential energy in the springs amounts to the force $\|v_i -m_i\|$ times the distance $\|v_i -m_i\|$.
\citet{levinson2020analysis} analyzed the gradient $\nabla_M L$ in the context of singularities. Their experiments showed that training with data being close to points with $\det (M)=0$, see \figref{fig:svd} right, increases the gradient magnitude, but does not notably impede training.

\paragraph{\GSO and the Procrustes problem}
As initially pointed out by \citet{bregier2021deep}, the \GSO representation \mbox{$r:=M=[\nu_1,\nu_2]\in\mathbb{R}^{3 \times 2}$} can be seen as a degenerate case of the Procrustes problem such that
\begin{equation} \label{eq:gso-procrustes-problem}
    \text{GSO}(M):=\underset{\epsilon \to 0^+}{\lim} \;  \underset{R\in \SO(3)}{\mathrm{arg\,min}}  \; \|\text{diag}(1, \epsilon, 0)\,R - M\|_{\mathrm{F}}.
\end{equation}
Thus, $\nu_1$ has by far the strongest influence on determining $R$, whereas $\nu_2$ merely determines the rotation around $\nu_1$. In principle, there are six different ways on how to perform the Gram Schmidt orthonormalization (around the axis $e_1$-$e_2$, $e_1$-$e_3$, $e_2$-$e_1$, $e_2$-$e_3$, $e_3$-$e_1$, $e_3$-$e_2$) which all place importance on different columns of the rotation matrix.
\GSO can be reduced to a 5D representation, which in practice seems to perform worse \cite{zhou2019continuity}.
\vspace{-1mm}

\paragraph{Comparing \GSO and \SVD}
Studies by \citet{zhou2019continuity, levinson2020analysis, bregier2021deep}, and our experiments confirm that \SVD often outperforms \GSO in rotation estimation. Why is that?

\citet{levinson2020analysis} attributes better performance of \SVD to the fact that Gaussian noise on $M$ results in twice the error in expectation in $R$ for \GSO compared to \SVD. 
\GSO predicts the first column of the rotation matrix by normalizing $\nu_1$. In turn noise on $\nu_1$ will directly affect the estimation of $R$, whereas for \SVD the effect of noise in $m_1$ also depends on $m_2$ and $m_3$. 
If we interpret the individual representation vectors as the output of different networks, then the SVD layer can be seen as an ensemble model architecture where all three ``networks'' equally contribute to the prediction whereas \GSO almost fully resorts to the first ``network''. 
It is a common strategy to deploy ensemble models rather than a single network as they often show increased \emph{robustness to input noise} and an unlucky initialization of network weights.

\begin{figure}
    \centering
        \vspace{-2.5mm}
    \includegraphics[width=\columnwidth]{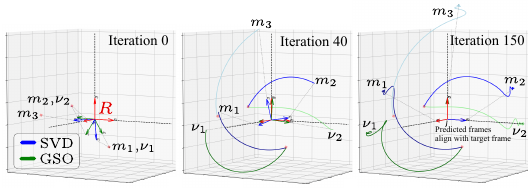}
    \vspace{-8mm}
    \caption{Illustration of how the representation vectors of \GSO $r:=[\nu_1,\nu_2]$ and \SVD $r:=[m_1,m_2,m_3]$ move through $\mathbb{R}^3$ during optimization. To find an $r$ that minimizes the loss \eqref{eq:gradientanalysis} to a target rotation $R:=[v_1,v_2,v_3]$, we perform 150 iterations of gradient-descent with momentum and plot the path of the representation vectors. \GSO can only minimize the loss, if $\nu_1$ aligns with $v_1$ while $\nu_2$ resides on the $v_1$-$v_2$-plane.} 
    \label{fig:feature_optimization}
    \vspace{-2mm}
\end{figure}

\citet{bregier2021deep} defined the loss function $L = v_1^T f(r) v_2$ with $v_i$ being vectors pointing in uniform random direction and $r$ denoting either quaternions, \GSO, or \SVD. For random representation vectors $r \sim \mathcal{N}(0,1)^n$, this loss function has a \emph{smaller absolute error to its linear approximation} using \SVD compared to using \GSO or quaternions. Inspired by \citet{bregier2021deep}, \cite{dinh2017sharp}, \cite{gilmer2021loss}, and \citet{lyle2023understanding}, we take a closer look at the \emph{gradients} of the loss function
\begin{equation} \label{eq:gradientanalysis}
    L(R, r) = \|\mathrm{vec}(R), f(r)\|,
\end{equation}
with the target rotation $R:=\mathbb{I}$ and $f(r)$ denoting either an \GSO or \SVD layer. Assume a network predicts $r$ such that its parameters $\theta$ could be updated via the gradient $\nabla_{\theta} r \circ \nabla_r L$. Then, we are interested in how $\nabla_r L$ varies when using either \GSO or \SVD. To illustrate this, we plot in \cref{fig:feature_optimization} the path of the representation vectors of \GSO and \SVD when optimizing \eqref{eq:gradientanalysis} via gradient-descent with momentum. 
Due to the way GSO and SVD construct $f(r)$, we observed instabilities in the optimization for $\nu_1 \approx e_3$, $\nu_1 || \nu_2$, $\text{det}(M)=0$, and $M=-R$.

\begin{figure}
    \centering
    \includegraphics[width=\columnwidth]{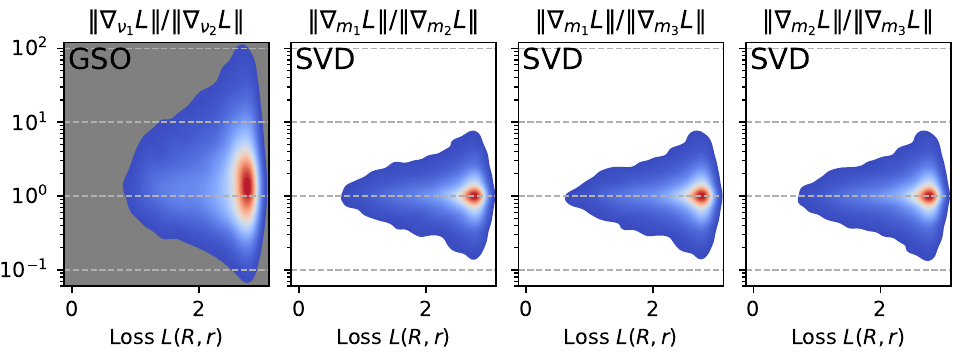}
    \vspace{-8mm}
    \caption{Density plot of the ratios between loss function gradients of \GSO and \SVD. The gradients of the loss \eqref{eq:gradientanalysis} are computed for 20000 randomly initialized $r$.}
    \label{fig:gradient_ratios}
    \vspace{-3mm}
\end{figure}

Moreover, the speed of convergence is notably affected by large ratios between the length of the representation vectors which particularly seems to affect \GSO. We compute the ratios between the length of the gradients of \eqref{eq:gradientanalysis} wrt.\ $\nu_1,\nu_2$ or $m_1,m_2,m_3$. As illustrated in \cref{fig:gradient_ratios} for 20000 uniform randomly initialized $r\in[-2,2]$, the \emph{gradient ratios remain notably closer to one} for \SVD compared to \GSO. Further, in \SVD, the distributions of gradient ratios are similar.
We conclude that in the hypercube with a side-length of four in which we sampled $r$, when using \GSO compared to \SVD considerably more instances of $r$ exhibit gradients that impede training.

Another argument why SVD performs better than GSO evolves around the \emph{higher-dimensionality} of the former. 
In various fields such as position encodings \cite{dufter2022positioninfo}, Koopman operator theory \cite{brunton2021modern}, and neural ODEs \cite{zhang2020approximation}, it has been observed that higher dimensional representations may benefit learning. This is particularly the case if the function we discern from data resides in a three-dimensional manifold such as $\SO(3)$.

\begin{mdframed}[backgroundcolor=gray!20]
\textbf{For rotation estimation:}
 use \SVD or \GSO.
If the regression targets are only small rotations, using quaternions with a halfspace-map is a good option.
\end{mdframed}

\paragraph{What about directly predicting the entries of $R$?} 
We find that this is worse than using Procruste-based methods.
We hypothesize that this is due to the properties of $\nabla_r L$.
When directly predicting rotation matrices with the Chordal distance, the network's predictions must exactly fit the entries of the rotation matrix to reduce the loss. Whereas in \SVD, the \emph{relative arrangement} of the representation vectors to each other determines $R$.

\subsection{What to be aware of for feature prediction?}
Feature prediction does not immediately suffer from double cover and a discontinuous $g$ because only the
forward function $f$ is part of the learning problem \ref{enum:feature-pred} and $f$ is typically well-behaved.
However, there are still problems due to the topological mismatch between $SO(3)$ and low-dimensional representations.
First, it is reasonable to employ a halfspace-map to avoid having less samples in each of the halfspaces that would need to be learned separately.

\begin{figure}[h]
    \vspace{-1mm}
    \centering
    \includegraphics[width=0.9\columnwidth]{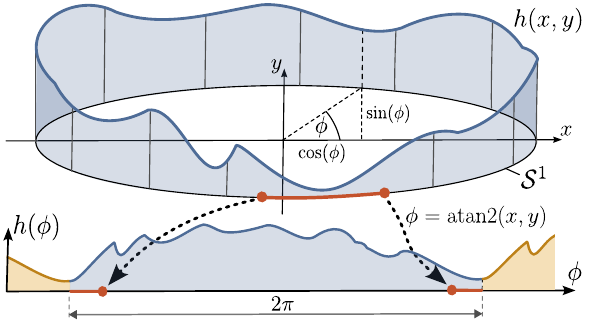}
    \vspace{-2mm}
    \caption{
    For $\SO(2)$, the function $\phi=f(\cos(\phi), \sin(\phi))$ is continuous, rendering feature prediction feasible. However, some connected sets in $\SO(2)$ are not connected in $\mathcal{R}$ (red). Data augmentation (orange) mitigates the effect of disconnected pre-images.
    }
    \label{fig:feature-prediction-illustration}
    \vspace{-4mm}
\end{figure}

\paragraph{Fixing low-dim representations using data augmentation:}
Low-dim representations and the halfspace map introduce a \emph{boundary} into the space (while $SO(N)$ has no boundary).
In \figref{fig:feature-prediction-illustration}, we illustrate this for $SO(2)$ for clarity.
Due to the broken cyclic boundary conditions, we have less samples at the boundary to generalize in these regions.
A simple trick is to use data-augmentation to create data beyond the boundary as shown in \figref{fig:quaternion_doublecover} and \ref{fig:feature-prediction-illustration}.

\paragraph{High-dimensional representations:}
Again, they work out of the box and perform better in practice, as we show in \cref{sec:learning}. More input dimensions allow for a simpler functional form for $h$.
A relation can be drawn to position encodings used in transformers \cite{dufter2022positioninfo}. A sin/cos representation (including higher frequencies) works better than using a scalar. 

\begin{mdframed}[backgroundcolor=gray!20]
\textbf{For feature prediction:}
 use \SVD or \GSO.
If under memory constraints,  quaternions with a halfspace-map and data-augmentation are viable.
\end{mdframed}

\section{Experiments} \label{sec:learning}
In this section, we empirically assess the earlier discussion and support the recommendations with various experiments on rotation estimation and feature prediction.

\begin{figure}
    \vspace{-2mm}
    \centering
    \includegraphics[width=.75\columnwidth, trim={-1cm 0 0 0}, clip]{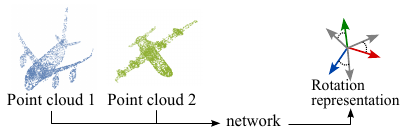}
    \includegraphics[width=0.95\columnwidth]{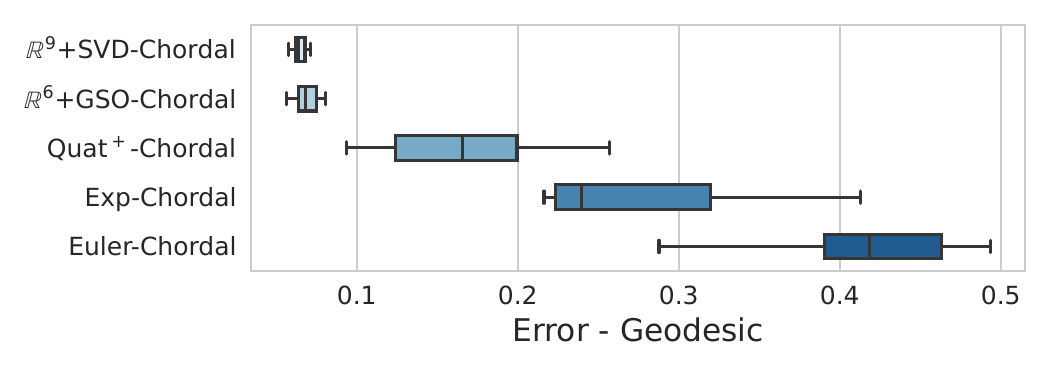}
        \vspace{-4mm}
    \caption{Point cloud alignment experiment.
    The best test data results from each representation are shown in which \SVD performs best. See \figref{fig:rotation_estimation_pcd:all} for additional results.
    }
    \label{fig:rotation_estimation_pcd}
    \vspace{-2mm}
\end{figure}

\vspace{-1mm}
\paragraph{Experiment 1 (rotation estimation): Rotation from point clouds}
The task is to predict from two point clouds $P_1, P_2 \in  \mathbb{R}^N$, with $N=3000$, the rotation $R$ such that $R P_2=P_1$ \cite{zhou2019continuity}.
We follow the model and dataset generation pipeline as \citet{levinson2020analysis}. Point clouds are extracted from a set of $726$ airplane CAD models and undergo uniformly sampled rotations in $SO(3)$ to eliminate any potential rotation bias in the dataset. For testing, 100 pairs are left out.
\cref{fig:rotation_estimation_pcd} shows that \SVD is best followed by \GSO.
\cref{fig:rotation_estimation_pcd:all} shows that for quaternions deploying a half-space map notably improves performance whereas distance picking reduces it.

\begin{figure}[t]
    \centering
    \includegraphics[width=0.85\columnwidth, clip, trim={0 0 -3mm 0}]{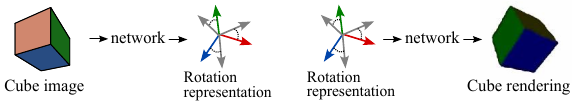}
    \includegraphics[width=0.95\columnwidth]{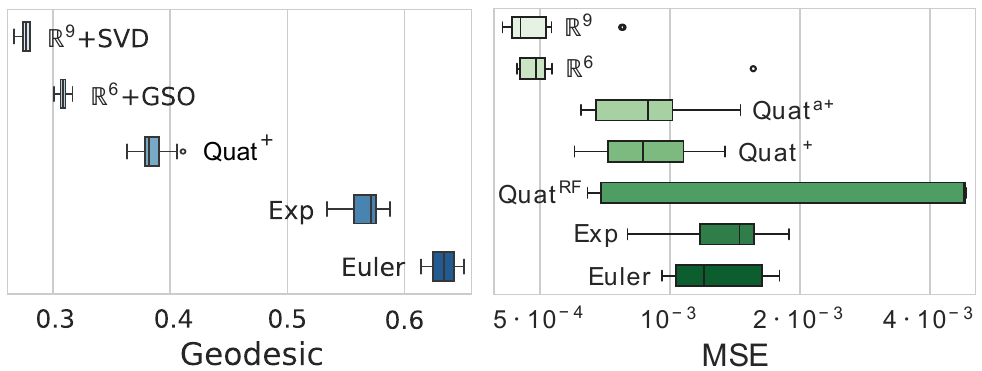}
    \vspace{-2mm}
    \caption{Experiment 2. Left: Rotation estimation from cube image where \SVD performs best. See \cref{fig:cube_experiment_img_to_pose_appendix} for additional results.
    Right: Network trained with pixel-MSE loss renders a cube from a given rotation. While quaternion augmentation slightly reduces the error, \SVD shows superior performance.
    }
    \label{fig:cube_experiment}
    \vspace{-3mm}
\end{figure}

\vspace{-1mm}
\paragraph{Experiment 2.1 (rotation estimation): Cube rotation from images}
In this experiment, we predict the orientation of a colorful cube represented as a rendered image.
The experimental setup is explained in \cref{sec:app:cube-exp}.
The results in \figref{fig:cube_experiment} (left) paint the the same picture as before showing that \SVD and \GSO perform notably better than low-dimensional representations.

\vspace{-1mm}
\paragraph{Experiment 2.2 (feature prediction): Cube rotation to images}
Now, we look at the inverse problem to Experiment 2.1 predicting the cube's image from an representation of its orientation. The experiment is further described in \cref{sec:app:cube-exp}. The results in \figref{fig:cube_experiment} (right) show that \SVD performed notably better than \GSO while quaternions benefit from deploying a half-space map.

\begin{figure}[t]
\vspace{-3mm}
    \centering
    \includegraphics[width=0.9\columnwidth]{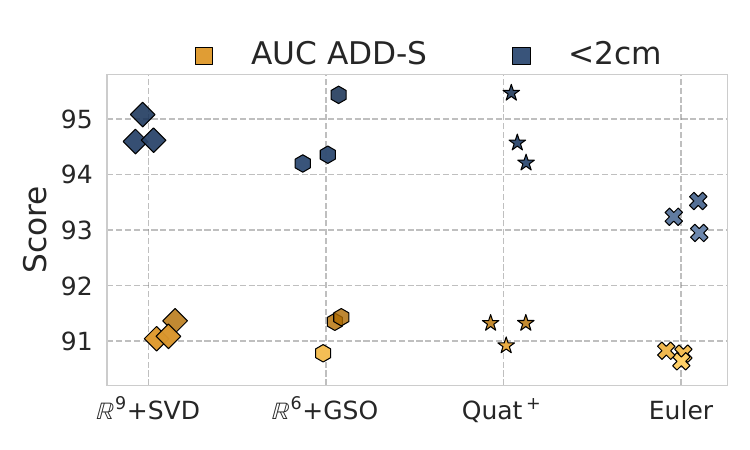}
    \vspace*{-4mm}
    \caption{6D-pose estimation from images on YCB-Video dataset. The metrics are averaged across the 21 objects. Euler angle perform worst while the other representations perform similarly.}
    \label{fig:6d-pose-est}
    \vspace{-3mm}
\end{figure}

\paragraph{Experiment 3 (rotation estimation): 6D object pose estimation from RGB-D images}
In this experiment, we re-train the network of \citet{wang2019densefusion} using different rotation representations. Simply put, the network predicts a translation and rotation representation which is used to transform an object's point cloud such that it aligns with the ground truth point cloud. This model has become a common baseline for object pose estimation from a single RGB-D camera image. \figref{fig:6d-pose-est} shows the average accuracy across all objects for three seeds. 
Euler angles perform notably worse than \SVD and \GSO whereas quaternions perform similarly. We hypothesise, that quaternions perform well due to the data set mostly containing small angles. For details, see \cref{sec:app:6D-pose-est}.

\begin{figure}[t]
    \centering
    \includegraphics[width=0.9\columnwidth, trim = 0 0 0 0, clip]{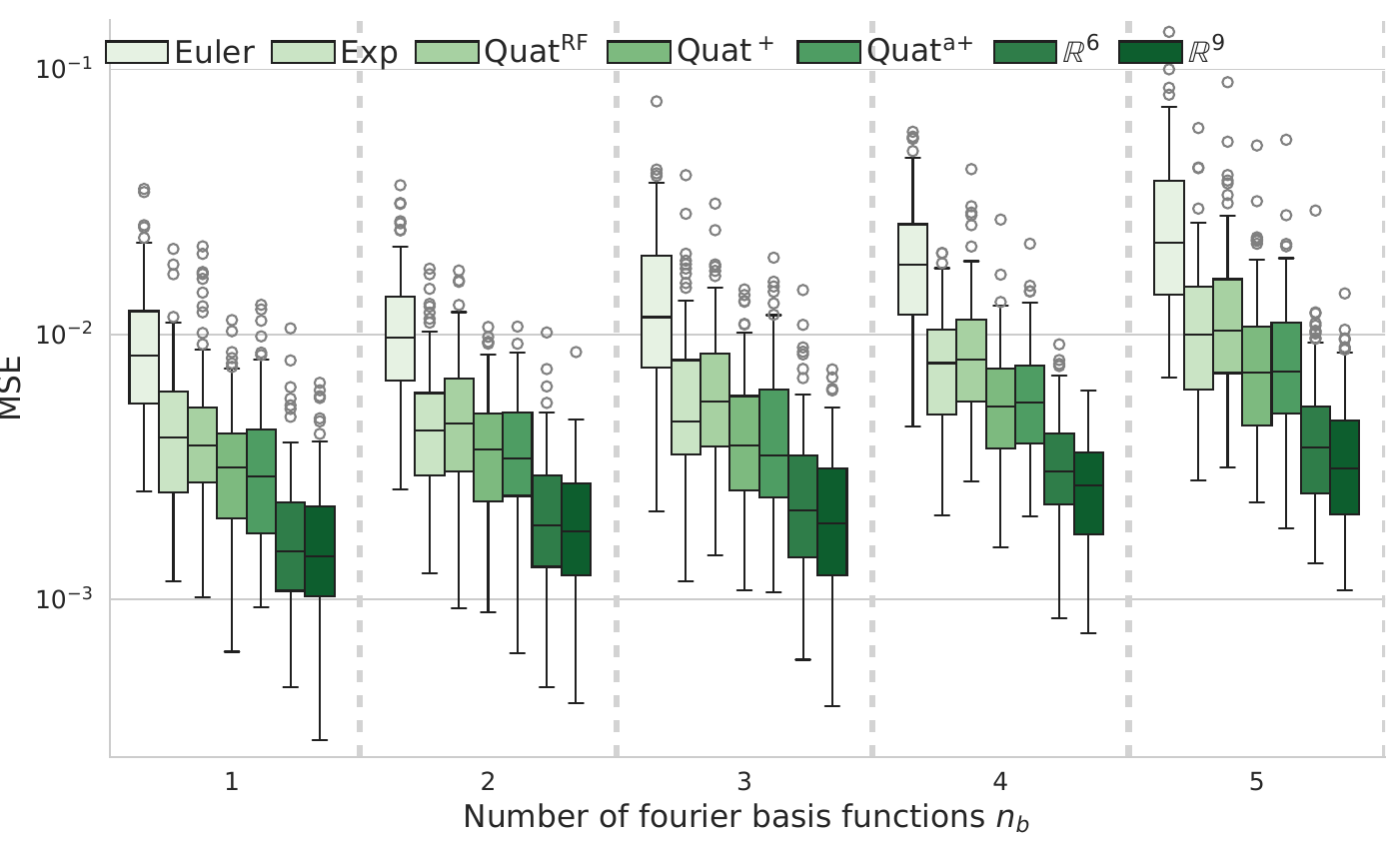}
    \vspace{-1mm}
    \caption{Feature prediction of a Fourier series that has 3D rotations as input. For increasing $n_b$, the distribution over test errors of 100 randomly sampled target functions is shown. Augmenting quaternions (Quat${}^{\text{a}\!+}$) notably improves prediction accuracy.}
    \label{fig:SO3_fourier_comparison}
    \vspace{-5mm}
\end{figure}

\vspace{-1mm}
\paragraph{Experiment 4 (feature prediction): $\SO(3)$ as input to Fourier series}
In this experiment, we analyze how function complexity affects feature prediction. The target function is chosen as
\begin{equation}
    \htrue(x) = \sum_{i=1}^{n_b} \left( A_k \cos\left(\frac{k\pi t(R)}{L}\right) + B_k \sin\left(\frac{k\pi t(R)}{L}\right) \right)
    \label{eq:fourier-target-function}
\end{equation}
with period $L=2$, fourier parameters $A_k, B_k$, and $t: \SO(3) \to \mathbb{R}$ being a 2-layer MLP with ReLu activations. Both the parameters of $t(\mathrm{vec}(R))$ and the Fourier parameters are randomly initialized. If we increase $n_b$, then the complexity of \eqref{eq:fourier-target-function} increases in expectation.
As further detailed in \secref{sec:SO3-fourier-series-appendix}, we approximate each randomly generated target function with an MLP which has a rotation representation as input. \figref{fig:SO3_fourier_comparison} shows the training results from $n_b=1$ up to $n_b=5$ for 100 target functions each. We keep the number of points fixed to 800, 200, and 1000 for the train, the validation, and the test data set, respectively. As we increase $n_b$, the target function becomes on average more wiggly and the model MSEs increase. Augmentation of quaternions (augmentation threshold b=0.1) has a notable effect for $n_b=1$ to $n_b=3$. Albeit, for larger $n_b$, the error arising from having no data points beyond the domain boundaries seem to be of minor importance.

\section{Conclusion}
In this work, we examine recent works on learning with rotations
and consider the case of rotation representations in the input (feature prediction) or in the output (rotation estimation) of a regression task.
We find it useful to consider the Lipschitz continuity of the function $g$ mapping from $\SO$(N) to the representation space $\mathcal R$.
We show how this mapping enters the learning problem and see why rotation representations with three or four parameters impede learning by inevitably introducing discontinuities into the target function, if the rotation is in the model's output.
We show that these discontinuities cannot be resolved by distance-picking, measuring distances in $\SO(3)$, or  half-space maps. In contrast, \SVD and \GSO stand out as superior choices when learning with rotations. In cases involving small angles, unit-quaternions properly mapped to a half space can be  practical for rotation estimation.

When rotations appear in the inputs of the regression task, discontinuities do not hinder learning. Nevertheless, high-dimensional representations are superior in practice.
When memory is constrained such that low-dimensional representations are preferred, the double cover property needs to be taken care of using half-space maps. Further, data augmentation reduces the impact of non-cyclic boundaries, such that viable performance improvements can be achieved.

Numerous extensions to the discussed topics are being proposed. Notably, \citet{peretroukhin_so3_2020} represents rotations through a symmetric matrix that defines an antipodally symmetric distribution over quaternions effectively mitigating the problems arising due to double cover. This approach outperformed \GSO on various benchmarks.
\citet{chen2022projective} notably improved prediction performance of \GSO, \SVD, and \cite{peretroukhin_so3_2020} layers whose predictions are fed into an L2 loss by adjusting the gradients during the backward pass which marginally increases computation times.

\subsection*{Acknowledgments}
The authors extend their gratitude to Shukrullo Nazirjonov whose curiosity on learning with rotations served as inspiration for this work. 

This work was supported by the ERC - 101045454 REAL-RL and the German Federal Ministry of Education and Research (BMBF) through the Tübingen AI Center (FKZ: 01IS18039B). Georg Martius is a member of the Machine Learning Cluster of Excellence, EXC number 2064/1 – Project number 390727645.

\subsection*{Impact Statement}
This paper presents work whose goal is to advance the field of Machine Learning. There are many potential societal consequences of our work, none which we feel must be specifically highlighted here.

\bibliography{main}

\newpage
\appendix
\onecolumn

\section{Notation}
Vectors are denoted as $a = [a_1, a_2, a_3] \in \mathbb{R}^{3 \times 1}$. The identity matrix of size three is denoted as $\mathbb{I}$. We denote the $n$ dimensional unit sphere as $\mathcal{S}^n=\{x\in\mathbb{R}^{n+1} : \|x\|=1\}$. Given a vector $a\in\mathbb{R}^d$, $\mathrm{diag}(a)\in\mathbb{R}^{d \times d}$ denotes a diagonal matrix with the entries of $a$ on its diagonal. 
Given a vector $a=[a_1,a_2,a_3]\in\mathbb{R}^3$, $a \times a = [a]_{\times} \cdot a$ where
\begin{equation}
    [a]_{\times} = \begin{bmatrix}0 & -a_3 & a_2 \\ a_3 & 0 & -a_1 \\ -a_2 & a_1 & 0 \end{bmatrix}.
\end{equation}
Given a matrix $M \in \mathbb{R}^{m \times n}$, the operation $\mathrm{vec}(M)$ rearranges the elements of $M$ as a vector such that $\mathrm{vec}(M)\in\mathbb{R}^{mn \times 1}$. Given a scalar $\phi \in \mathbb{R}$, its absolute value is denoted as $|\phi| \geq 0$.

\section{Matrix representation of SO(3)}
$\SO(3)$ denotes the \emph{special orthogonal group}. This name comes from the fact that all rotation matrices are special ($det(M)=1$) and orthogonal.
In three dimensions, every rotation is described by a rotation matrix $M\in \SO(3)$ being a real-valued matrix of the form
\begin{equation}
M=\begin{bmatrix}
\mid & \mid & \mid \\
m_{1} & m_{2} & m_{3} \\
\mid & \mid & \mid
\end{bmatrix} \in \mathbb{R}^{3 \times 3}
\end{equation}
whose three column-vectors $m_1, m_2, m_3$ span a Cartesian frame such that
\begin{equation}
\|m_i\|=1 \text{ for } i=1,2,3, \: \: m_3 = m_1 \times m_2. \label{eq:rot_constraints}
\end{equation}

\begin{figure}[htb]
    \centering
    \includegraphics[width=0.49\textwidth]{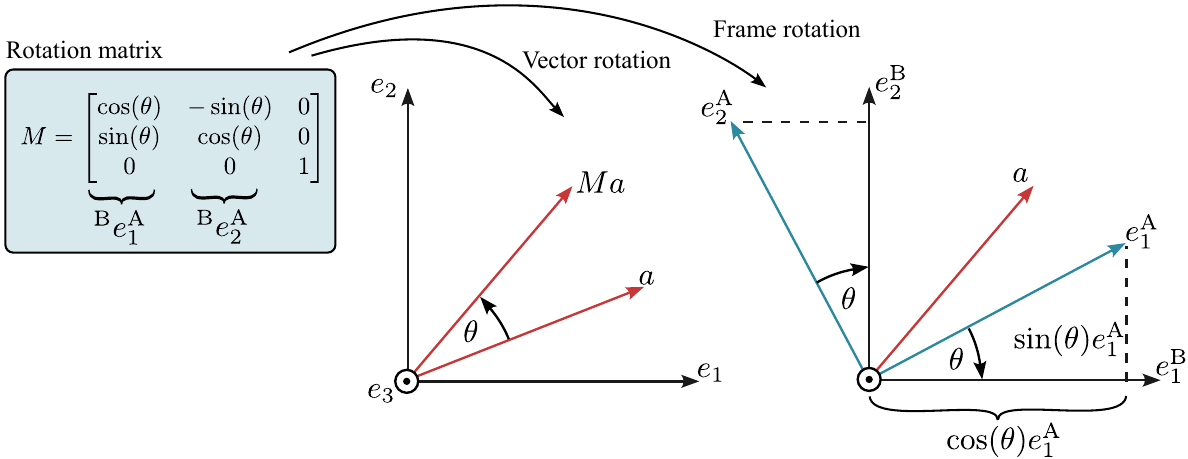}
    \vspace{-3mm}
    \caption{A rotation matrix $M \in SO(3)$ rotates a vector $a \in \mathbb{R}^3$ to $Ma \in \mathbb{R}^3$. Alternatively, the multiplication of a frame`s unit vectors with $M$ shows that the columns of the rotation matrix denote the frame's unit vectors after rotation.}
    \label{fig:perspective_rotations}
\end{figure}

As illustrated in Fig.\,\ref{fig:perspective_rotations}, rotations are commonly represented by a vector $a \in \mathbb{R}^3$ transformed by matrix $M$ to produce $b = M a \in \mathbb{R}^3$.
Alternatively, we gain geometric intuition on the rotation matrix itself if we imagine the vector $a$ to be described wrt.\ to a frame $\{A\}$, writing ${}^{A}a=[a_1, a_2, a_3]$, and the vector ${}^{B}a$ being obtained as
\begin{equation}
{}^{B} a=M\, {}^{A} a = m_1 a_1 + m_2 a_2 + m_3 a_3
\label{eq:rot-basis}
\end{equation}
where $\{B\}$ denotes a frame rotated relative to $\{A\}$.
In turn, we can think of the rotation matrix in terms of its columns $m_1:={}^{\text{B}}e_1^{\text{A}}, m_2:={}^{\text{B}}e_2^{\text{A}}, m_3:={}^{\text{B}}e_3^{\text{A}}$ being the unit vectors $e_i$ of $\{A\}$ expressed relative to a rotated frame $\{B\}$. Simply put, the columns of a rotation matrix span a Cartesian frame relative to another Cartesian frame.

\begin{figure}
    \centering
    \includegraphics[width=1.\columnwidth, trim={0.5cm, 0, 0cm, 0}, clip]{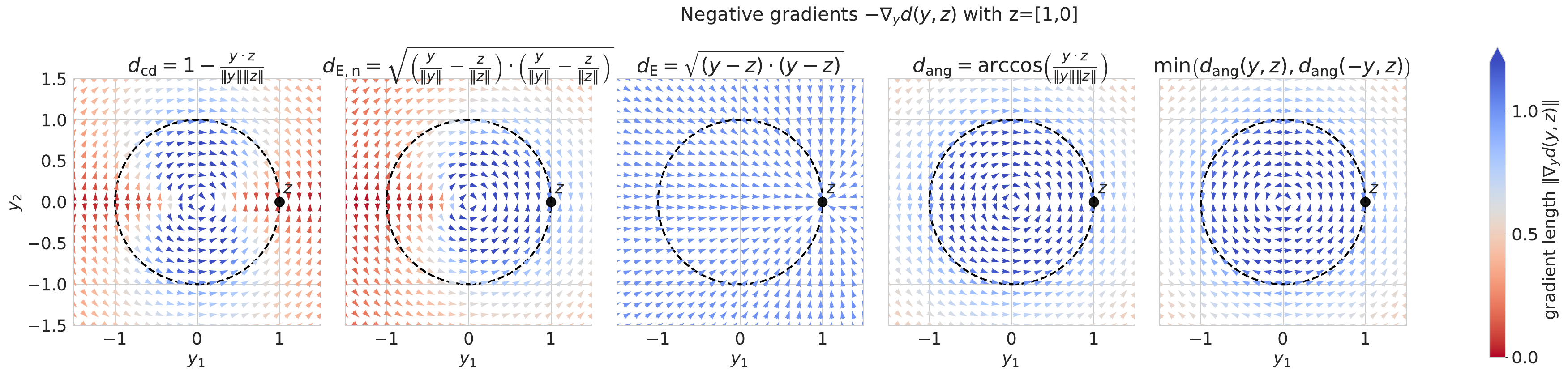}
    \vspace{-4mm}
    \caption{
    Given a fixed target vector $z=[1,0] \in \mathcal{S}^1$, the negative gradients $-\nabla_{y} d(y,z)$ are plotted for different $y=[y_1,y_2] \in\mathbb{R}^2$. For the Euclidean distance with input normalization and cosine distance, the gradient $\nabla_{y} d(y,z)$ is zero when $y$ is antipodal to $z$. For the Euclidean distance the gradients point from $z$ to $y$. For the angular distance, the gradients are tangential to the unit-circle. When training a model $y=h(a,\theta)$, we use $\nabla_{y} d(y,z)$ to adjust $\theta$ such that $y$ is close to the target vector $z$.}
    \label{fig:distance_gradients}
\end{figure}

\section{Intuition on some terminology from linear algebra}

\textbf{``Orthogonal''} refers to $MM^T=M^TM=I$ such that the columns of the matrix are orthogonal to each other and have unit length.

The \textbf{dot product} between two vectors $a \cdot b=\|a\|\|b\| \cos(\theta)$  ($\theta$ is the angle between $a$ and $b$) is zero if the vectors are orthogonal, one if they are colinear unit vectors pointing in the same direction, and -1 if they are colinear unit vectors pointing in opposite directions. If $a$ and $b$ are unit vectors, then the inner product corresponds to $\cos(\theta)$.

The \textbf{determinant} of a 3D matrix measures the signed volume of the parallelepiped spanned by the column vectors of the matrix.

\textbf{``Special''} refers to $\det (M) = 1$.
 If $\det(M)=1$, then the column vectors of $M$ form a right-hand coordinate system. Orthogonal matrices with determinant of one are called \textbf{proper rotations}, whereas orthogonal matrices with determinant of negative one are called \textbf{improper rotations}. Improper rotations combine rotation with reflection, the product between two improper rotations is a proper rotation. We encounter improper rotations in the description of rotations via geometric algebra where rotations are described by means of two reflections.

\section{Additional aspects on learning with rotations}

\subsection{Why learn with rotation representations that have three or four parameters?} \label{sec:why-low-dim}

\paragraph{Computational efficiency} A representation with $3$ instead of $9$ parameters may significantly reduce memory consumption. When advocating for low-dimensional rotation representations, it is tempting to resort to the ``curse of dimensionality'' which relates to the volume of the space between data points growing exponentially with increasing data dimensionality. However, $\SO(3)$ rotations reside on a three-dimensional manifold in $\mathbb{R}^9$ such that it is unclear if the curse of dimensionality aggravates learning for higher-dimensional rotation representations.

\paragraph{Forward dynamics} If a network predicts the forward dynamics of a multi-body system, small errors in the prediction of the body's rotation over time may accumulate. In turn, the predicted state eventually does not remain in the proximity of the three-dimensional manifold in $\mathbb{R}^9$ on which the training data has been collected. \citet{makinen2008rotation} provides comprehensive theoretical analysis on finite changes in rotations.

\begin{figure}[t]
    \centering
    \begin{picture}(100,100)
    \put(0,100){Kinematics}
    \put(0,90){of a human}
    \put(0,80){skeleton}
    \put(50,0){\includegraphics[clip, trim=8cm 2cm 8cm 3cm]{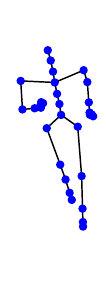}}
    \end{picture}
    \hspace{1cm}
    \includegraphics[width=0.5\textwidth]{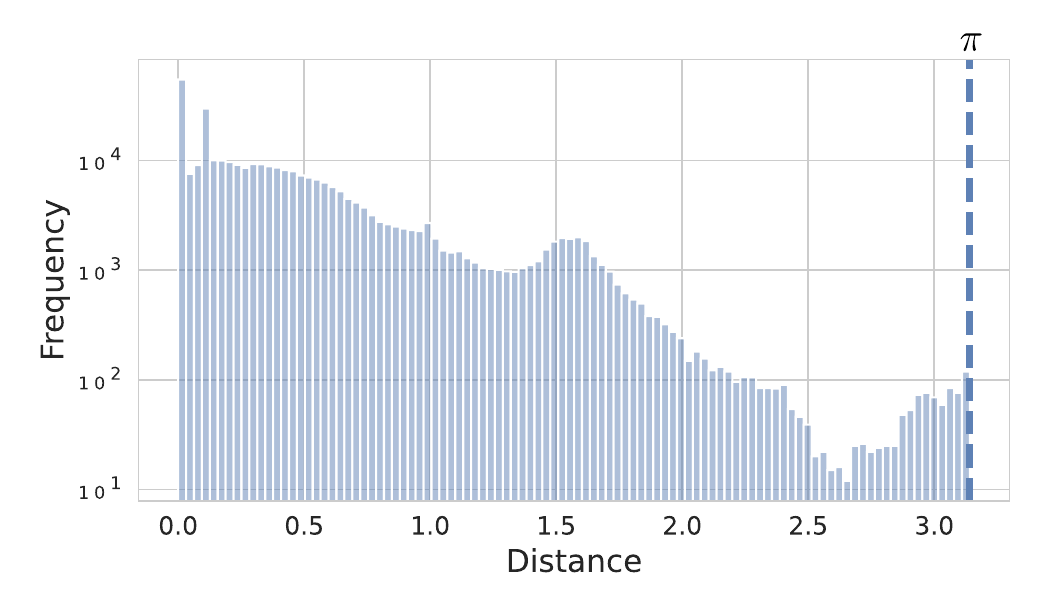}
    \vspace{-0.5cm}
    \caption{\textbf{Left:} Illustration of a human skeleton as contained in the CMU MoCap database \cite{cmu-database}. The database contains marker position of human skeletons and the corresponding rotations between the lines. \textbf{Right:} Distribution of the distances between the identity rotation and the rotations in the CMU MoCap database \cite{cmu-database}. This data base is used in the rotation estimation experiments of \citet{zhou2019continuity, levinson2020analysis, bregier2021deep, pepe2022learning}. In $\SO(3)$, two rotations are maximally $\pi$ away from each other when using the Geodesic distance. 
    In this data set, most rotations are close to the identity rotation such that the singularities inherent in low-dimensional rotation representations may not significantly affect rotation estimation.}
    \label{fig:IK-exp}
\end{figure}
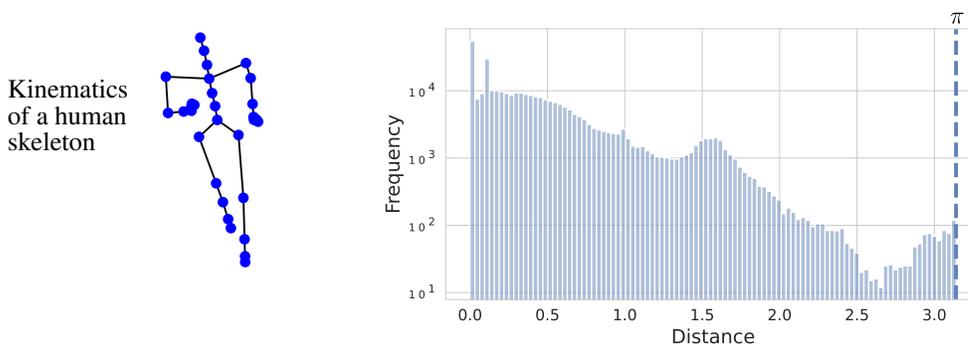

\subsection{How do we augment quaternions for feature prediction?} \label{sec:quaternion-augmentation}
The following code performs data augmentation of a data set containing unit-quaternions to improve learning with rotations in the model's output:
\begin{verbatim}
x_quat = x_rot.as_quat(canoncial=True)
epsilon = 0.1
condition = x_quat[:, -1] < epsilon  
x_quat = np.r_[-1 * x_quat[condition,:], x_quat]
y_features = np.r_[y_features[condition,:], y_features]
\end{verbatim}
Assume that each row of the array \verb|x_quat|$\in \mathbb{R}^{m \times 4}$ denotes a unit-quaternion. Here, we follow the convention in SciPy that the last dimension of the quaternion is its scalar part. To every quaternion a feature is being associated which is stored in the array \verb|y_features|$\in \mathbb{R}^{m \times n}$. As shown below, we first ensure that all quaternions lie on the half unit sphere such that $q_w \geq 0$ \eg by using ScipPy's rotation library \verb|x_rot.as_quat(canoncial=True)|. Then we check which quaternions have a scalar part $q_w<$\verb|epsilon| and store the indices of these quaternions in an array. These quaternions are then mapped to the other half-sphere by multiplication with $-1$ and concatenated to the data set. \verb|y_features| is also augmented using the same indices. 

\paragraph{Batch-wise augmentation of quaternions}
In the experiment implementation,  instead of following the above approach for augmenting quaternions, we augment solely the rotations in the current batch via the PyTorch code:
\begin{verbatim}
x_quat[torch.logical_and(torch.rand(x_quat.size(0)) < 0.5, 
                         x_quat[:, 3] < 0.1)] *= -1
\end{verbatim}
Before running the above augmentation, we ensured that \verb|x_quat| is a torch array of size $m \times 4$ that only contains canonical quaternions where $m$ denotes the batch size. As this approach only operates on the current batch it is more memory-efficient but computationally more costly. Further, with batch-wise augmentation a quaternion $q$ and its complement $-q$ cannot simultaneously occur in the augmented batch if $q$ appears only once in the data.

\subsection{Are rotors underrated?} \label{sec:rotors}
\citet{pepe2022learning} contends that using bi-vectors (3D) and rotors (4D) for rotation estimation might rival rotation representations employing the Gram-Schmidt orthonormalization (6D). However, bi-vectors / rotors are in fact computationally equivalent to exponential coordinates / quaternions as shown in a detailed code comparison by \citet{marc2020code}. While a geometric algebra perspective on rotations (including rotors and bi-vectors) may be more intuitive than using quaternions, as exemplified by \citet[p.\,87-91]{macdonald2010linear} and \citet{marc2020remove}, we remain cautious about expecting significant performance differences for rotation estimation.

Additionally, while Table 2 in \cite{pepe2022learning} suggests that networks can learn the map from rotors to bi-vectors, these representations are not exempt from the singularities in $g(R)$ causing $h(x)$ to become discontinuous during rotation estimation. Further, \citet{pepe2022learning} observed that rotors obtained using the Caley-transform performed differently than rotors obtained using the matrix-exponential. We hypothesize that one of these maps performs a half-space map while the other does not.

We advocate for further analysis that considers the proximity of representation vectors to singularities, measures all errors in $\SO(3)$ rather than in rotor space, and tries using a half-space map for quaternions and exponential coordinates. One experiment found in \citet{zhou2019continuity, levinson2020analysis, bregier2021deep, pepe2022learning} is ``Inverse Kinematics with CMU MoCap data'' (see \secref{sec:kienamtics-experiment-app}), for which we analysed how far the rotations are from the unit rotation. We hypothesized that for some joints the anatomy of human kinematics does not support large angle ranges and in turn the data set contains mostly small rotations (recall that a rotation $R \in \SO(3)$ is small if $\|\mathbb{I}-R\|_{\mathrm{F}}\leq\sqrt{2}$). Indeed, as shown in \cref{fig:IK-exp}, this data set contains significantly more small rotations that are close to the unit rotation which reduce the effect of singularities on learning with rotation representations with three or four dimensions. 

\begin{figure}[t]
    \centering
    \includegraphics[width=0.9\textwidth]{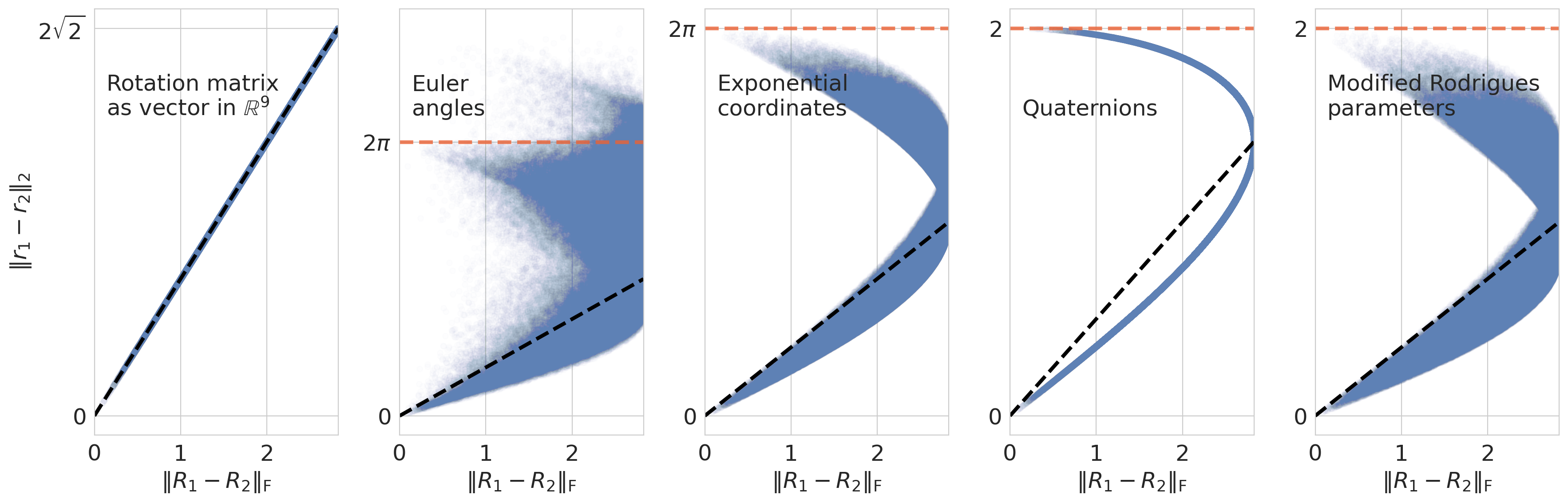}
    \caption{Chordal distances $\|R_1 -R_2\|_{\mathrm{F}}$ between randomly sampled $R_1,R_2\in\SO(3)$ and distance $\|r_1-r_2\|$ between corresponding rotation representations $r_1=g(R_1),r_2=g(R_2)$. The full width of $\mathcal{R}$ is marked by red lines. Ideally, the ratios between the distances, aka the Lipschitz constant of $g(R)$, is close to the black lines whose slope amounts to the ratio between the minimum width of $\mathcal{R}$ and $\SO(3)$. For Euler angles how much a distance in $\SO(3)$ is altered to a distance in $\mathcal{R}$ strongly depends on where $R_1,R_2$ lie in $\SO(3)$. In comparison, for unit-quaternions, distances in $\SO(3)$ of a certain magnitude of any $R_1,R_2$ correspond to similar distances in $\mathcal{R}$.}
    \label{fig:lipschitz-extended}
\end{figure}

\subsection{How does distance picking and computing distances in $\SO(3)$ affect rotation estimation?} \label{sec:app-distance-picking}
The term distance picking relates to using metrics such as \eqref{eq:distance-picking-1} and \eqref{eq:distance-picking-2} for rotation estimation. As pointed out by \cite{huynh2009metrics}, these functions are pseudo-metrics in $\mathcal{R}$ but metrics in $\SO(3)$. How do the prediction of a neural network $h(x,\theta)$ look like that is trained using such a metric? At the beginning of training the network's parameter are randomly initialized and the output of the network points in different direction of $\mathcal{R}$. Then during training the networks parameters are being updated to move the network's output close to the rotation representation or their negative complement that is closest according to \eg \eqref{eq:distance-picking-1}. Now for simplicity's sake, assume that the training data consists of two features $a_1, a_2$ which reside close to each other such that $a_1= a_2 + \epsilon$. Yet, due to an unlucky initialization $q \approx h(a_1,\theta)$ and $-q \approx h(a_2,\theta)$. In this case, the optimizer deploying \eqref{eq:distance-picking-1} adjusts the network's parameters to move $\hat q_1 = h(a_1,\theta)$ close to $q$ and $\hat q_2 = h(a_2,\theta)$ close to $-q$. In turn, the network's ability to interpolate between features that map to similar rotations is severely hindered while the network's gradients may vary significantly in magnitude.

\subsection{Additional aspects on learning with rotations}
Besides the continuity of $g(a)$, numerous other aspects affect learning with rotations.  In particular, the distribution of the training data and observability of the object should be analyzed prior to training. 

If only a small part of the rotation space is present within the training dataset, one cannot generalize to the full rotation space. This problem has been commonly experienced by researchers training models on the YCB-Video dataset where most objects are biased toward standing upright. Similarly, the CMU MoCap database \cite{cmu-database} contains mostly small angle ranges.

The second aspect is, the observability of the task at hand. If objects are (partially)-occluded or symmetrical, the learning problem may be ill-posed as pointed out by \cite{saxena2009learning}.

\subsection{How does the ratio between distances induced by $g(r)$ look like for axis-angles and exponential coordinates?}
\cref{fig:lipschitz-extended} shows also the ratio between the Chordal distance $\|R_1-R_2\|_{\mathrm{F}}$ of randomly sampled rotation matrices $R_1,R_2 \in \SO(3)$ relative to the corresponding distance in $\mathcal{R}$ both for exponential coordinates and axis-angles. For Euler angles the ratio between the distances strongly depends on where the rotation matrices are in $\SO(3)$. For exponential coordinates and modified Rodrigues parameters this distance ratio looks similar to unit-quaternions yet is clearly not as neatly aligned. Modified Rodrigues parameters $[\tilde \omega, \bar \alpha]$ are obtained by taking the angle $\alpha$ in the axis angle representation and transforming it to $\bar \alpha = \tan(\alpha / 4)$. Modified Rodrigues parameters are bijectively mapped to quaternions via stereographic projection as succinctly illustrated in \cite{terzakis2018modified}.

\subsection{How does \SVD and \GSO compare in terms of computation times?}
\begin{figure}[h]
        \centering
        \includegraphics[width=0.5\textwidth]{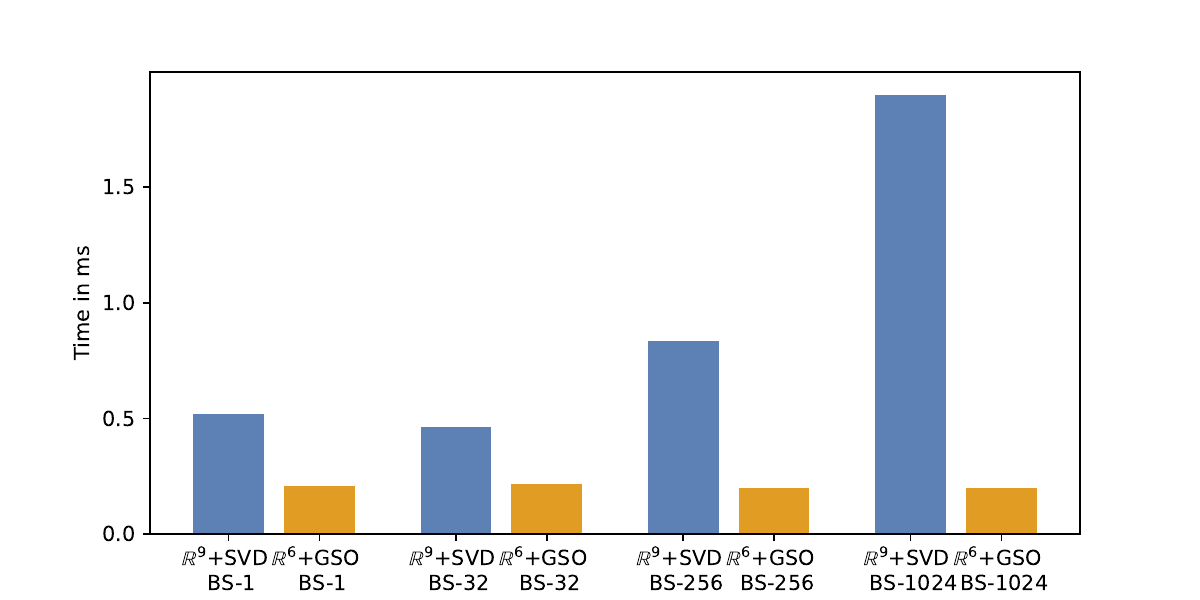}
        \caption{Experiment 2.2: Timing comparison of \GSO and \SVD to $\SO(3)$ mapping for different batch sizes $[1,32,256,1024]$ on a Nvidia RTX3060.}
        \label{fig:gso_svd_timing}
\end{figure}
We previously, concluded that for most applications, one should use the continuous rotation representation. 
In the following, we analyze the additional computational cost required for training and inference when using the \ensuremath{\mathbb{R}^6+ GSO} and \ensuremath{\mathbb{R}^9+SVD} rotation representation. 
We timed the forward pass for 100 samples of the GSO and SVD mapping operations within Experiment 2.1. 
We followed the best practices and took into account the asynchronies execution on the GPU, and warm-started the accelerator by inferencing 100 samples before timing the mapping functions. 
All numbers are reported using an Nvidia RTX3060.

The results can be seen in \cref{fig:gso_svd_timing}. 
The SVD is computationally more expensive than the GSO. 
The GPU-based \ensuremath{\mathbb{R}^9+SVD} takes around \SI{0.5}{ms} for a batch size of 1. 
However, there exist multiple CPU-optimized SVD implementations that can decrease the inference time for small batch size (\SI{.002}{ms} \url{https://github.com/ericjang/svd3}), or one can resign to using Jax for better efficiency. 
When looking at a batch size of 1024, the SVD only takes \SI{2}{ms}. 
Specifically crucial for deploying neural networks is the increase in latency for small batch sizes (\eg running an object pose estimator on a robot), which is nearly neglectable when using a CPU-based SVD implementation. 
Another critical aspect is the reduced throughput during training which, when using high-capacity neural network architectures like CNNs, Transformers, or Diffusion Models, is neglectable as well (\eg \SI{200}{ms} ResNet152 vs \SI{2}{ms} SVD). 
However, specifically when training very small networks (\eg 2-layer MLP), it can be beneficial to directly optimize on \ensuremath{\mathbb{R}^9} (without SVD) and without GSO leading to only a minor performance decrease (see \cref{fig:cube_experiment_img_to_pose_appendix}). 

In summary,we conclude that for most applications, the overhead in terms of compute when using \ensuremath{\mathbb{R}^9+SVD} 
 compared to \ensuremath{\mathbb{R}^6+ GSO} is small.
We highlight the clear trade-off between performance, increased latency, and throughput that one has to take into consideration. 

\section{Further experiment details}
The project code is available at: \href{https://github.com/martius-lab/hitchhiking-rotations}{github.com/martius-lab/hitchhiking-rotations}

\subsection{Details of Experiment 1 (Rotation estimation): Rotation from point clouds} \label{sec:point-cloud-appendix}

The dataset creation consists of first sampling 3000 points from the 726 airplane CAD models provided by the ModelNet dataset and secondly rotating them with rotations uniformly sampled from $SO(3)$, using "SciPy" \cite{2020SciPy-NMeth}. The 726 pairs of rotated point clouds and their rotations are split into 626 training and 100 testing samples. To get the different rotation representations, the rotations are transformed into Euler angles (xyz-intrinsic rotations), canonical quaternions (mapped to have positive scalar), and exponential coordinates. For \SVD and \GSO the matrix form is used, and the representation is captured by a mapping function as explained below.

The architecture of the model used for learning follows exactly the architecture described in \citet{levinson2020analysis} and \citet{zhou2019continuity}. First, point clouds are embedded with a simplified PointNet (4 MLP layers 64, 128, 256, 1024) followed by a global max-pooling. This is then regressed via two LeakyReLu-activated MLP(512) layers with dropout followed by an MLP(d) which outputs the d-dimensional vector of the corresponding representation.

During training, we minimize either the mean absolute error (MAE) or mean squared error (MSE). For quaternions, we also experiment with distance picking (-dp) as training loss. The \SVD and GSO representation are trained by projecting the predicted vector $\mathbb R^d$ to $\mathbb R^9$ and minimizing the chosen loss there. We incorporate early stopping in our training process with a patience of 10 epochs and a maximum of 100 epochs to ensure a fair and effective training of diverse representations.

\subsection{Details of Experiment 2: Cube image to\,/\,from rotation} \label{sec:app:cube-exp}

\begin{figure}[t]
    \centering
    \includegraphics[width=1\textwidth]{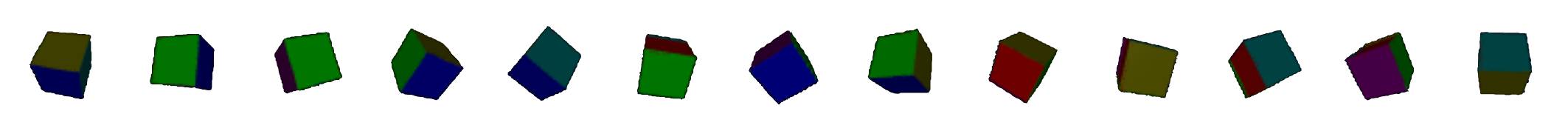}
    \vspace{-0.8cm}
    \caption{Experiment 2.2: Cube Rendering Examples.}
    \label{fig:cube_rendering_examples}
\end{figure}

We create a dataset of differently rotated cubes at the same positions with respect to a fixed camera. The dataset is generated once and kept fixed across training seeds. 
The training, validation, and test dataset consists of 2048 image/rotation pairs.
The rotations, used to render the images are sampled uniformly distributed for both datasets.
The images are rendered using the MuJoCo simulator \cite{todorov2012mujoco} at a resolution of $64 \times 64$.
In all experiments, we use the Adam optimizer and train for up to $1000$ epochs. 
We early stop the training if no improvement is achieved over 10 validation epochs. 
We report all metrics across 10 random seeds, for the best-performing validation model on the test dataset. 
All hyperparameters are constant across the same experiment.
For the orientation prediction task, we use a learning rate of $0.001$ and a batch size of $32$. The network is given by a 3-layer MLP, with input size $64*64*3$, hidden dimension $[256, 256]$, ReLU activation functions, and the output size is determined by the dimension of the rotation representation.

Within the main paper (Experiment 2.1, \cref{fig:cube_experiment}), we use the geodesic distance as a training objective and report the Chordal distance. 
In \cref{fig:cube_experiment_img_to_pose_appendix}, we report the geodesic distance as well as the Chrodal distance and compare different rotation representations and training objectives. 

When training on the Chordal or geodesic distance, we at first map the predicted representation to \ensuremath{\mathbb{R}^9}.
The MAE and MSE are directly computed between the predicted rotation representation and the target rotation, without mapping to \ensuremath{\mathbb{R}^9}. 
All metrics for \ensuremath{\mathbb{R}^9} can be optimized with and without SVD. 
For \ensuremath{\mathbb{R}^6}, only the MAE and MSE loss can be optimized without applying GSO.
For quaternions, we additionally train on the cosine distance (CD) and use MSE distance picking (MSE-DP).

Generally, we found that the continuous representations strongly outperform quaternion, exponential coordinate, and Euler angle representations. 
One of the key findings is, that by applying an output mapping to the rotation representation and using a metric defined on \ensuremath{\mathbb{R}^9} (e.g, the Chordal or geodesic distance), one cannot overcome the limitations intrinsic to the output representation of the network. 

For the quaternion representation, we observe that first mapping to \ensuremath{\mathbb{R}^9} and using the geodesic distance and the Chordal distance leads to an overall better performance compared to cosine distance (CD) and applying distance picking on the MSE (MSE-DP).
One would hypothesize when evaluating the Chordal distance, one should optimize for the Chordal distance during training. However, for \ensuremath{\mathbb{R}^9}, \ensuremath{\mathbb{R}^6} optimizing for the geodesic distance leads to better performance. We would recommend the reader experimentally try out both objective functions.

For the feature prediction/image generation task, we use a learning rate of $0.01$ and a batch size of $128$. The network is given by a Convolutional Neural Network (CNN), consisting of one LinearLayer and 4 blocks of ConvTranspose2d, BatchNorm2d and ReLU activation function. 
The loss function and evaluation metric is given by the pixel-wise MSE.

\subsection{Details of Experiment 3 (Rotation estimation): 6D object pose estimation from RGB-D images} \label{sec:app:6D-pose-est}
\begin{table*}[]
\caption{Experiment 3: Per-object 6D pose estimation results, measured by Area under the Curve of the ADD-S (AUC) and percentage of predictions with a ADD-S lower the \SI{2}{cm} (<\SI{2}{cm}) on the YCB-Video dataset. The averages are provided in the last row. The asterisk '*' indicates objects with symmetry.} \label{tab:dense_fusion_per_object}
\vspace{3mm}
\centering
\tiny
\begin{tabular}{@{}lccp{0.001cm}ccp{0.001cm}ccp{0.001cm}cc}\toprule
 & \multicolumn{2}{c}{\textbf{Euler}} && \multicolumn{2}{c}{\textbf{$\mathbb{R}^6$+GSO}} & &\multicolumn{2}{c}{\textbf{$\mathbb{R}^9$+SVD}} && \multicolumn{2}{c}{\textbf{Quat per-pixel}}   \\
 \cmidrule{2-3} \cmidrule{5-6} \cmidrule{8-9} \cmidrule{11-12}
 & \multicolumn{1}{r}{<\SI{2}{cm}} & \multicolumn{1}{c}{AUC} && \multicolumn{1}{c}{<\SI{2}{cm}} & \multicolumn{1}{c}{AUC} &&\multicolumn{1}{c}{<\SI{2}{cm}} & \multicolumn{1}{c}{AUC} && \multicolumn{1}{c}{<\SI{2}{cm}} & \multicolumn{1}{c}{AUC}  \\
 \midrule
 
\textbf{Master chef can}   & 100.0 \pmm 0.0  & 95.0 \pmm0.2  &&  100.0 \pmm0.0  &   94.9 \pmm0.39  &&    100.0 \pmm0.0   &    95.1 \pmm 0.04  &&   100.0 \pmm 0.0  &   95.3 \pmm 0.14        \\
\textbf{Cracker box}       & 99.2 \pmm 0.46  & 92.1 \pmm 0.41 &&  98.2 \pmm 1.75  &   93.3 \pmm 0.2   &&    99.0 \pmm 0.29   &    92.4 \pmm 0.12  &&   99.4 \pmm 0.3   &   92.8 \pmm 0.31   \\
\textbf{Sugar box}         & 99.8 \pmm 0.16  & 95.2 \pmm 0.09 &&  100.0 \pmm 0.0  &   95.4 \pmm 0.15  &&    100.0 \pmm 0.04  &    95.4 \pmm 0.26  &&   100.0 \pmm 0.0  &   95.4 \pmm 0.28   \\
\textbf{Tomato soup can}   & 96.9 \pmm 0.0   & 93.6 \pmm 0.23 &&  96.9 \pmm 0.0   &   93.8 \pmm 0.13  &&    96.9 \pmm 0.0    &    93.7 \pmm 0.04  &&   96.9 \pmm 0.0   &   93.9 \pmm 0.12   \\
\textbf{Mustard bottle}    & 100.0 \pmm 0.0  & 95.6 \pmm 0.71 &&  99.8 \pmm 0.26  &   95.6 \pmm 0.52  &&    100.0 \pmm 0.0   &    95.8 \pmm 0.17  &&   100.0 \pmm 0.0  &   96.0 \pmm 0.2    \\
\textbf{Tuna can}          & 100.0 \pmm 0.0  & 95.4 \pmm 0.45 &&  100.0 \pmm 0.0  &   95.7 \pmm 0.08  &&    100.0 \pmm 0.0   &    95.9 \pmm 0.26  &&   100.0 \pmm 0.0  &   95.2 \pmm 0.41   \\
\textbf{Pudding box}       & 100.0 \pmm 0.0  & 94.1 \pmm 0.06 &&  100.0 \pmm 0.0  &   93.7 \pmm 0.32  &&    99.8 \pmm 0.22   &    93.7 \pmm 0.68  &&   99.8 \pmm 0.22  &   94.1 \pmm 0.55   \\
\textbf{Gelatin box}       & 100.0 \pmm 0.0  & 96.7 \pmm 0.26 &&  100.0 \pmm 0.0  &   97.2 \pmm 0.33  &&    100.0 \pmm 0.0   &    97.0 \pmm 0.25  &&   100.0 \pmm 0.0  &   97.0 \pmm 0.2    \\
\textbf{Potted meat can}   & 92.5 \pmm 0.12  & 89.7 \pmm 0.09 &&  92.9 \pmm 0.12  &   89.6 \pmm 0.46  &&    93.0 \pmm 0.16   &    89.8 \pmm 0.11  &&   92.8 \pmm 0.18  &   89.9 \pmm 0.47   \\
\textbf{Banana}            & 88.2 \pmm 7.13  & 90.0 \pmm 1.39 &&  93.2 \pmm 2.2   &   92.2 \pmm 0.82  &&    93.9 \pmm 1.63   &    92.3 \pmm 0.41  &&   91.5 \pmm 2.48  &   91.2 \pmm 0.91   \\
\textbf{Pitcher}           & 100.0 \pmm 0.0  & 93.3 \pmm 0.2  &&  99.4 \pmm 0.83  &   93.3 \pmm 0.46  &&    99.9 \pmm 0.08   &    93.5 \pmm 0.53  &&   100.0 \pmm 0.0  &   93.6 \pmm 0.57   \\
\textbf{Bleach}            & 98.4 \pmm 2.18  & 94.1 \pmm 0.97 &&  99.9 \pmm 0.08  &   94.4 \pmm 0.23  &&    100.0 \pmm 0.0   &    94.8 \pmm 0.07  &&   99.8 \pmm 0.16  &   94.4 \pmm 0.05   \\
\textbf{Bowl*}             & 57.9 \pmm 1.72  & 84.8 \pmm 1.22 &&  69.1 \pmm 18.48 &   85.3 \pmm 0.7   &&    70.4 \pmm 11.26  &    85.5 \pmm 0.38  &&   71.1 \pmm 10.77 &   86.4 \pmm 0.27   \\
\textbf{Mug}               & 99.7 \pmm 0.37  & 94.7 \pmm 0.17 &&  100.0 \pmm 0.0  &   94.7 \pmm 0.66  &&    100.0 \pmm 0.0   &    95.2 \pmm 0.18  &&   100.0 \pmm 0.0  &   95.4 \pmm 0.22   \\
\textbf{Power drill}       & 95.0 \pmm 3.39  & 92.2 \pmm 1.15 &&  96.8 \pmm 3.61  &   92.9 \pmm 1.11  &&    96.2 \pmm 1.52   &    91.8 \pmm 0.47  &&   96.5 \pmm 1.33  &   92.6 \pmm 0.2    \\
\textbf{Wood block*}       & 94.2 \pmm 6.19  & 87.8 \pmm 1.41 &&  98.5 \pmm 0.78  &   87.4 \pmm 1.16  &&    95.5 \pmm 1.79   &    87.5 \pmm 0.71  &&   94.4 \pmm 3.45  &   87.7 \pmm 1.24   \\
\textbf{Scissors}          & 98.7 \pmm 1.14  & 94.4 \pmm 0.98 &&  96.5 \pmm 3.42  &   92.8 \pmm 2.07  &&    97.6 \pmm 2.31   &    92.2 \pmm 2.01  &&   98.5 \pmm 0.69  &   92.5 \pmm 1.67   \\
\textbf{Marker}            & 90.4 \pmm 3.79  & 93.4 \pmm 0.52 &&  99.1 \pmm 0.33  &   94.5 \pmm 0.09  &&    99.7 \pmm 0.26   &    94.7 \pmm 0.34  &&   98.4 \pmm 1.46  &   94.7 \pmm 0.43   \\
\textbf{Clamp*}            & 77.2 \pmm 0.13  & 72.6 \pmm 0.53 &&  76.5 \pmm 0.63  &   72.2 \pmm 0.43  &&    77.1 \pmm 0.75   &    71.7 \pmm 0.63  &&   77.9 \pmm 0.07  &   71.3 \pmm 0.21   \\
\textbf{Large clamp*}      & 69.7 \pmm 0.84  & 69.0 \pmm 0.17 &&  71.1 \pmm 0.78  &   72.2 \pmm 1.68  &&    70.7 \pmm 0.37   &    73.6 \pmm 0.2   &&   72.8 \pmm 0.85  &   72.6 \pmm 1.76   \\
\textbf{Foam brick*}       & 100.0 \pmm 0.0  & 92.2 \pmm 0.17 &&  100.0 \pmm 0.0  &   93.6 \pmm 0.54  &&    100.0 \pmm 0.0   &    92.9 \pmm 0.25  &&   100.0 \pmm 0.0  &   92.9 \pmm 0.65   \\

 \hline
\textbf{All objects} & 93.3 \pmm 0.15  & 90.8 \pmm 0.07 && \textbf{94.7} \pmm 0.08 & \textbf{91.2} \pmm 0.04 && 94.6 \pmm 0.24 &  91.1 \pmm 0.16 && \textbf{ 94.7} \pmm 0.63 & \textbf{91.2} \pmm 0.20 \\ \bottomrule

\end{tabular}

\end{table*}

We use the implementation provided of \citet{wang2019densefusion} and test the performance of the network on the YCB-Video Dataset consisting of 21 different objects and follow the evaluation of \citet{xiang2018posecnn}, reporting the Area under the Curve of the ADD-S (AUC) and the percentage of objects with an ADD-S smaller than \SI{2}{cm} (<\SI{2}{cm}). We only consider the initial pose estimation without refinement. Therefore, we adapt the \emph{per-pixel} method while solely altering the network output between different rotation representations. We use a total of 3 seeds because of the high training cost. 
\tabref{tab:dense_fusion_per_object} reports the mean and standard deviation.

\subsection{Details of Experiment 4 (Feature prediction): $\SO(3)$ as input to Fourier series} \label{sec:SO3-fourier-series-appendix}
The network is given by a 3-layer MLP with input size being determined by the rotation representation, hidden dimension $[256, 256]$, ReLU activation functions, and the output size of one. The networks were trained for 400 Epochs using Adam (standard settings in PyTorch with a starting learning rate of $0.001$). At each iteration a batch of 64 data points was drawn from 800 train data points and validated on 200 validation points. After training, the best model with respect to the validation loss was used to compute a test loss on 1000 data points. For train, validation, and test loss we used the RMSE.

\subsection{Experiment 5 (Rotation estimation): Inverse Kinematics with CMU MoCap data} \label{sec:kienamtics-experiment-app}
The works of \citet{zhou2019continuity, levinson2020analysis, bregier2021deep, pepe2022learning} do rotation estimation for an Inverse Kinematics problem. In this setup, the model estimates joint rotations from the joint positions of a human pose as shown in \cref{fig:IK-exp}. These rotations are to be estimated relative to a canonical “t-pose”, defined as a skeleton in a standing position with arms stretched. We refrained from reproducing this experiment as the dataset mostly contains angles close to the unit rotation as shown in \cref{fig:IK-exp}. This corresponds to the “small-angle” case as discussed in Section \ref{sec:4.1}.

This plot was generated using 10,000 samples from the CMU MoCap database \cite{cmu-database}. The sampling process involved filtering out videos with 100 frames (static poses), randomly selecting 760 videos, and subsampling them to 30 frames each. From this collection of human poses, we uniformly sampled 10,000 poses. The joint rotations for each frame were then extracted relative to parent joints and expressed in relation to the “t-pose” joint rotations.

\begin{figure}
    \centering
    \includegraphics[width=\textwidth]{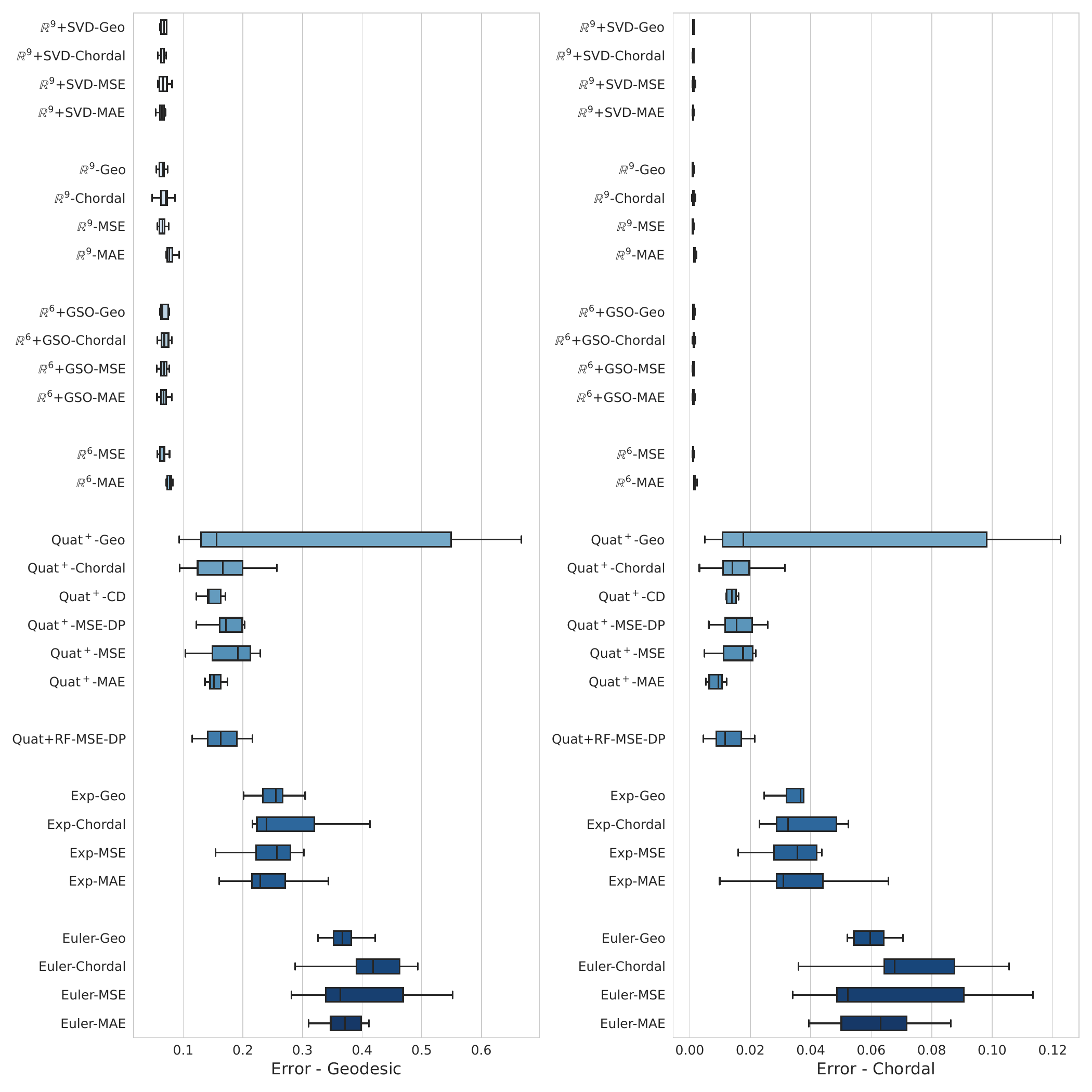}
    \vspace{-0.2cm}
    \caption{Experiment 1: Point clouds to rotation.
    Each box shows the test errors (Geodesic distance or Chordal distance) of ten networks trained on a specific rotation representation and loss function. 
    For quaternions, the cosine distance '-CD' as in \eqref{eq:cosine_distance} and distance picking metric '-MSE-DP' as in \eqref{eq:distance-picking-1} have also been used. "Quat-RF" refers to using quaternions which have been randomly multiplied by either -1 or 1. We only show "Quat-RF-MSE-DP" as using "Quat-RF" with CD, MAE, and MSE yielded errors four to seven times larger as the shown results. 
    $\mathbb{R}^6$ and $\mathbb{R}^9$ refers to directly using the first two and three column vectors of the rotation matrix, respectively. 
    Notably, resorting to the MAE can improve results for low-dimensional representations, as observed in the instances of quaternions, Euler angles, and exponential coordinates.
     }
    \label{fig:rotation_estimation_pcd:all}
\end{figure}

\begin{figure}[t]
    \centering
    \includegraphics[width=0.95\textwidth]{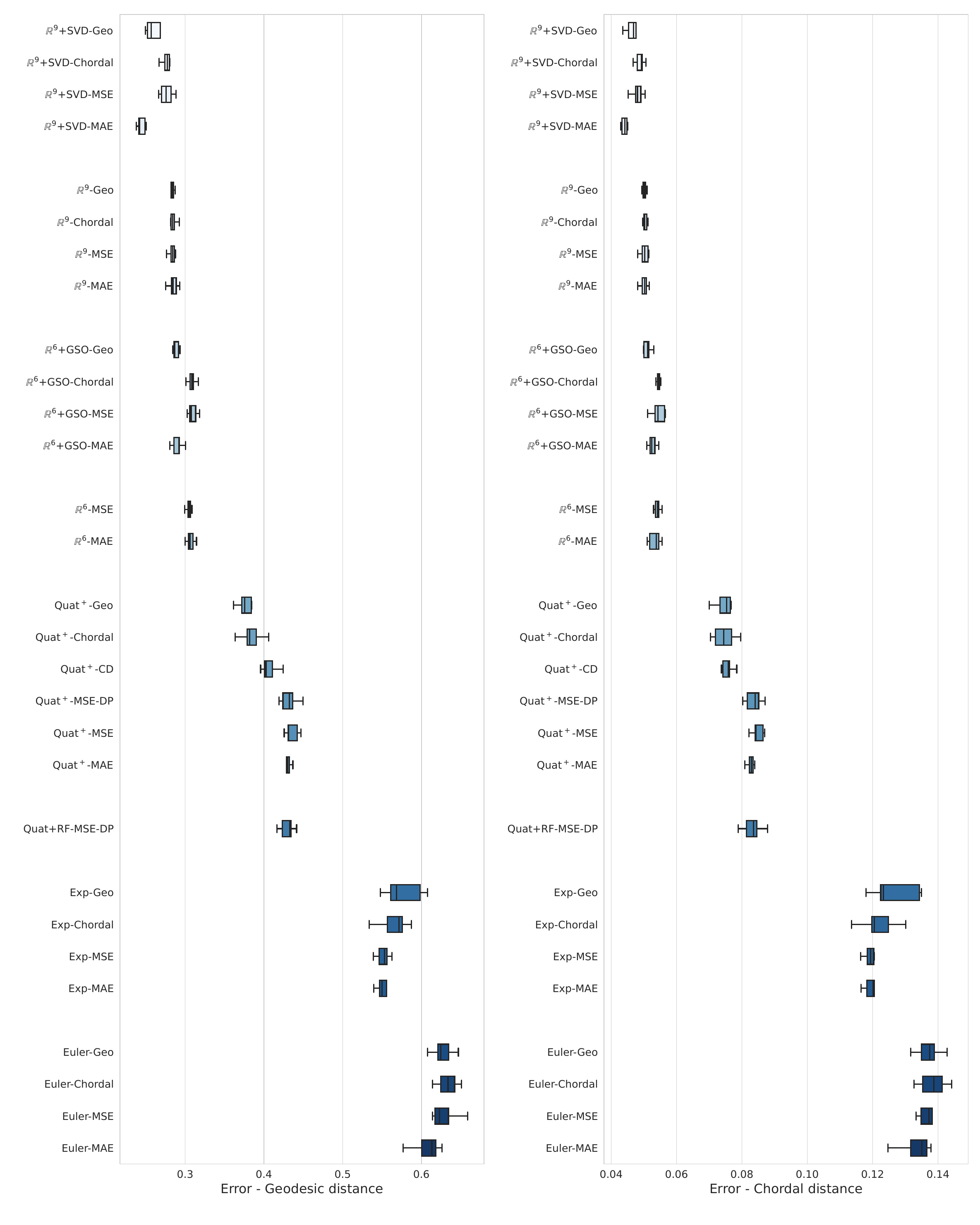}
    \vspace{-0.2cm}
    \caption{Experiment 2.1: Cube image to rotation. Each box shows the test errors (Geodesic distance or Chordal distance) of ten networks trained on a specific rotation representation and loss function. For quaternions, the cosine distance '-CD' as in \eqref{eq:cosine_distance} and distance picking metric '-MSE-DP' as in \eqref{eq:distance-picking-1} have also been used. The observed relative performance aligns with previous experiments illustrated in \cref{fig:rotation_estimation_pcd:all}.
    "Quat-RF" refers to using quaternions which have been randomly multiplied by either -1 or 1. We only show "Quat-RF-MSE-DP" as using "Quat-RF" with CD, MAE, and MSE yielded errors four to six times larger as the shown results. $\mathbb{R}^6$ and $\mathbb{R}^9$ refers to directly using the first two and three column vectors of the rotation matrix, respectively.
    }
    \label{fig:cube_experiment_img_to_pose_appendix}
\end{figure}


\end{document}